\documentclass{article}

\usepackage{arxiv}

\usepackage[utf8]{inputenc} 
\usepackage[T1]{fontenc}    
\usepackage{hyperref}       
\usepackage{url}            
\usepackage{booktabs}       
\usepackage{amsfonts}       
\usepackage{amsmath}
\usepackage{amssymb}
\usepackage{chemformula}
\usepackage{nicefrac}       
\usepackage{microtype}      
\usepackage{cleveref}       
\usepackage{lipsum}         
\usepackage{tikz}
\usetikzlibrary{positioning,fit,arrows.meta}
\usetikzlibrary{calc}
\usepackage{graphicx}
\usepackage{natbib}
\usepackage{doi}
\usepackage{subcaption} 
\usepackage{multicol,multirow}  
\usepackage{makecell}
\usepackage{algorithm}
\usepackage{algorithmic}
\def\argmax{\displaystyle\operatornamewithlimits{arg\,max}}

\title{Crystal Structure Prediction using Graph Neural Combinatorial Optimization }

% ...

\renewcommand{\shorttitle}{\textit{arXiv} Template}

\newif\ifuniqueAffiliation
\uniqueAffiliationtrue

\usepackage{authblk}

\author[1]{%
	{\hspace{1mm}Stavros~Gerolymatos}%
}
\author[2,3]{%
	{\hspace{1mm}J. Kyle~ Brubaker}%
}
\author[2]{%
	{\hspace{1mm}Martin J. A.~ Schuetz}%
}
\author[1]{%
	{\hspace{1mm}Vladimir~V.~Gusev}%
}

\affil[1]{School of Computer Science and Informatics, University of Liverpool, Liverpool, England, UK}
\affil[2]{Amazon Advanced Solutions Lab, Seattle, Washington 98170, USA}
\affil[3]{Arm Inc., San Jose, California 95134}
\affil[ ]{{\{s.gerolymatos, vladimir.gusev \}@liverpool.ac.uk}, maschuet@amazon.com, kyle.brubaker@arm.com}

\begin{document}

\maketitle

\begin{abstract}
Crystalline materials are widely used in technological applications, yet their discovery remains a significant challenge. As their properties are driven by structure, crystal structure prediction (CSP) methods play a central role in computational approaches aiming to accelerate this process. Previously, CSP has been approached from a combinatorial optimization perspective, with the core challenge of allocating atoms on a fine grid of predefined discrete positions within a unit cell while minimizing their interaction energy. Exact mathematical optimization methods provide guaranteed solutions, but they become computationally expensive for large-scale instances, where the atomic configuration space grows rapidly, particularly in the absence of additional symmetry constraints. In this work, we introduce a neural combinatorial optimization approach to the atom allocation challenge and, subsequently, CSP, based on graph neural networks (GNNs), which can effectively sample from the distribution of feasible structures in an unsupervised manner. We leverage expander graphs to construct computational graphs over discrete positions that capture both short- and long-range interactions between atoms, and employ the Gumbel–Sinkhorn approach to enforce the desired stoichiometry of the generated structures. We demonstrate that our method outperforms classical heuristic approaches and is competitive with a commercial optimization solver across a range of chemical compositions. This enables the use of ever-expanding GPU infrastructure to tackle the inherent combinatorial challenges of CSP, paving the way for scaling beyond current capabilities.
\end{abstract}


\section{Introduction}
The discovery of novel functional materials has enabled technological innovation in areas such as energy storage~\cite{Zhao2020}, catalysis~\cite{Zhao2019} and semiconductor design~\cite{Sumida2012}. At the same time,  the search for new materials remains a challenging endeavor requiring extensive experimentation and human intuition. In order to accelerate the discovery process a range of computational tools has emerged taking advantage of high-throughput screening~\cite{Curtarolo2013}, public materials databases~\cite{10.1063/1.4812323},~\cite{Kirklin2015},  and a wide range of machine learning models~\cite{Chen2019GraphNetworks} to identify promising candidates for synthesis. 

Crystal structure prediction (CSP) is a central component of computational discovery methods in inorganic chemistry, as structure determines the stability and properties of a material~\cite{Oganov2019},~\cite{Collins2017, article}. At a high level, CSP aims to determine the positions of atoms in space that minimize their interaction energy.  Although CSP approaches can be used to finetune the structures of already known materials using template based allocation, their greatest potential lies in the identification of novel structural types that have not been previously reported in databases~\cite{doi:10.1021/acs.accounts.4c00694}. The quantum nature of interatomic interactions, as well as the computational cost of solving atomic allocation problems, pose inherent obstacles to the widespread use of CSP~\cite{adamson1}.

In a recent breakthrough, CSP was approached from the combinatorial optimization (CO) point of view~\cite{gusev2023optimality}. CO is a field of optimization that aims to compute the minimum or maximum of an objective function, typically expressed in terms of a set of integer variables, over a finite set of feasible solutions subject to some constraints. CO problems have a wide range of critical real-world applications, among which are drug therapy planning~\cite{Pulkkinen2020CancerSelective}, vehicle routing/scheduling~\cite{10.5555/2190621}, logistics management ~\cite{paschos},  and structure search~\cite{Yin2022}, and have had a profound impact on advancing theoretical computer science and discrete mathematics over the last few decades~\cite{Kar72},~\cite{kuhn}. To deal with these often NP-hard and non-convex problems, researchers suggested exact (such as Gurobi~\cite{gurobi}), approximate~\cite{vazirani} and heuristic ~\cite{metaheur} solvers. The recent emergence of massive parallel computational devices such as GPUs~\cite{Shen2025} has led to the development of data-dependent and learning-based methods such as~\cite{BENGIO2021405},~\cite{gasse} to solve CO problems. These methods take advantage of some specific patterns or characteristics found in the given CO instances and aim to rapidly compute (approximate) solutions for them.  In this paper, we introduce a learning-based method that can leverage GPU infrastructure to address the inherent combinatorial challenge of CSP. 

Graphs are fundamental in the study of CO because they serve as models for many real-world phenomena, with the Traveling Salesman Problem (TSP) being one of the most well-known examples.In materials science, graphs are commonly used to represent molecules and crystal structures~\cite{cgcnn}. Graph neural networks (GNNs)~\cite{10.1109/TNN.2008.2005605} are well-established deep learning architectures that are used with graph-structured inputs and are trained to solve tasks such as node/graph classification and link prediction.~\cite{pignn} blended concepts from physics, optimization theory and GNNs to efficiently solve hard CO problems without supervision. In another unsupervised approach,~\cite{10.5555/3495724.3496283} combined GNNs with neural probabilistic sampling to compute low-cost valid solutions to CO tasks. Towards the same direction,~\cite{karalias2025geometric} introduced a geometric self-supervised method based on convex geometry and Carathéodory’s theorem. Finally, a notable practical application of GNNs in CO corresponds to chip placement, as studied by~\cite{chip}.

GNNs iteratively propagate information between neighboring nodes using the message-passing neural network (MPNN) framework ~\cite{messagepassing}: node representations are computed by aggregating vector-based messages from their neighbors defined by the (computational) graph with their own representations. In practice, however, for many tasks it is essential to consider interactions between nodes that are not directly connected. Although stacking more GNN layers offers a straightforward way to achieve that, it has been found to lead to over-smoothing~\cite{rusch2023surveyoversmoothinggraphneural},~\cite{oono2020graph},~\cite{zhao2020pairnorm}, a main shortcoming of the MPNN framework referring to the phenomenon in which node representations tend to become increasingly similar as the number of GNN layers grows. Another shortcoming of MPNNs corresponds to over-squashing~\cite{oversquashing}, the phenomenon in which messages between distant nodes are exchanged with significant loss of information, due to some graph connectivity properties, known as bottlenecks. Over-squashing affects the expressivity and representation power of a GNN and can be addressed by decoupling the computational graph used for message-passing from the problem-defined graph. As a result, the flow of information between distant nodes for graphs that are not friendly to message-passing can be improved. Different approaches to do this include graph rewiring, where the graph connectivity is modified through edge removal and addition to reduce the over-squashing effect~\cite{topping2022understandingoversquashingbottlenecksgraphs}, the introduction of virtual (artificial) nodes that are connected to existing nodes to enable long-distance message propagation~\cite{10.5555/3737916.3738806} and the use of expander graphs. Expander graphs~\cite{deac2022expandergraphpropagation},~\cite{shirzad2023exphormer} have been identified as appealing because they are constructed to eliminate over-squashing and reduce bottlenecks by allowing effective communication between distant nodes within a small number of hops.

\begin{figure}
    \centering
    \includegraphics[width=1\linewidth]{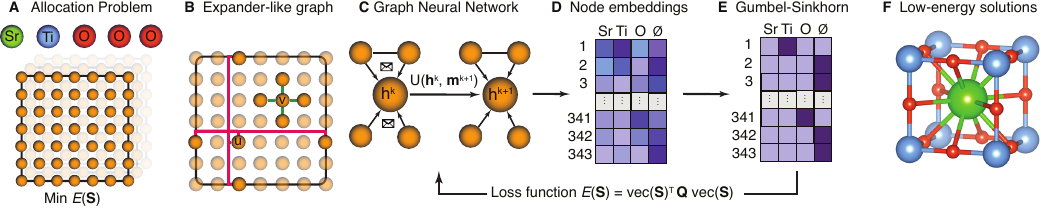}
\caption{\textbf{GNT-CSP: neural combinatorial approach to CSP. A.} Following~\cite{gusev2023optimality}, a dense grid of points is defined within a unit cell as potential atomic positions. We seek to identify the optimal, in terms of energy, allocation of the required number of atoms corresponding to a chosen stoichiometry (in this figure, \ch{SrTiO3}), to this set of positions.
\textbf{B.} A computational graph based on a classic expander graph construction is built over these positions. It comprises both short-range edges (green) and long-range edges (red).
\textbf{C.} The aim is to classify each node of this graph to the most appropriate atom for allocation (\ch{Sr}, \ch{Ti}, \ch{O}, or void $\varnothing$ in this figure), in  an unsupervised manner. Message passing takes place, and the resulting node embeddings (\textbf{D.}) are used to solve a node classification problem that involves minimizing the total interaction energy. \textbf{E.} The Gumbel--Sinkhorn approach is applied to the final node embeddings before backpropagation to ensure that the predicted structure satisfies the desired stoichiometry. \textbf{F.} Predicted structure corresponding to \ch{SrTiO3}.}
    \label{fig:schematic_figure}
\end{figure}

In this work, we introduce a neural combinatorial approach for CSP, which we refer to as Graph Neural Transportation (GNT-CSP).  Following~\cite{gusev2023optimality}, we seek a minimum energy allocation of atoms to a uniform grid of points within a given unit cell. Two types of constraints are always present in this setting: (a) no two atoms are allowed to occupy the same position; (b) the desired stoichiometry dictates the number of atoms and their species. Our GNT-CSP approach, inspired by~\cite{pignn}, inherently encodes both CSP constraints, eliminating the need for explicit penalty terms, and finds high-quality feasible structures without relying on any solutions precomputed by a solver. Starting from a grid of points that discretizes the unit cell and defines potential atomic positions, we construct a 3D graph based on the Gabber-Galil graph. We then train a GNN to compute node representations that are projected into the feasible solution space and sample high-quality solutions minimizing the total interaction energy. In essence, we solve the CSP problem as a multiclass node classification problem, where each node can be assigned to either an atom of a specific chemical element or void (the position remains unoccupied), while ensuring that the CSP constraints are satisfied and the solution space of atom allocations is explored to compute feasible low-energy atom allocations. A schematic diagram illustrating the key steps of our proposed method is provided in Figure~\ref{fig:schematic_figure}.

Our contributions are as follows.
\begin{itemize}
    \item We introduce a GNN-based approach with which we can inherently handle both CSP constraints and obtain low-energy solutions by projecting the node embeddings into the feasible solution space.
    \item We introduce a 3D graph topology, inspired by the Gabber-Galil graph that combines both local and long-range connections, providing a more appropriate option for message-passing than simple local cutoff graphs.
    \item We empirically demonstrate that our approach can find high-quality solutions to combinatorial CSP for a variety of compositions.
\end{itemize}

\section{Related Work}

We begin with a brief overview of the existing literature for CO problems using machine learning. For a detailed review of CO with GNNs, we refer the reader to~\cite{10.5555/3648699.3648829}. Further, we provide a summary of related methods for solving the CSP task. 

\textbf{ML in Combinatorial Optimization}. 
Over the last few years, ML algorithms have emerged as a novel paradigm towards solving CO problems. The development of machine learning algorithms has had a significant impact on the solving of combinatorial optimization problems. Supervised learning models ~\cite{selsam2019learningsatsolversinglebit}, ~\cite{nair2021solvingmixedintegerprograms} are trained with labeled data computed by numerical solvers. Because of this, these models often require pre-solving many hard problems, which can take time and can be computationally expensive. Supervised learning models can also struggle to generalize across different problem size scales. These limitations appear in the context of material discovery, where collecting labeled training data often requires computationally heavy and time-extensive experiments. 

Unsupervised learning models are introduced to account for these limitations by jointly optimizing the objective of a CO's problem and a measure of constraints' violation, without requiring labeled data in terms of precomputed solutions.~\cite{Schuetz_2022} formulated the NP-hard graph coloring problem as a multiclass node classification problem. In addition to this, they introduced a physics-inspired GNN to solve it by minimizing an unsupervised loss function based on the statistical Potts model \cite{RevModPhys.54.235}.

Alternatively, generative approaches such as diffusion models ~\cite{diffusion} and generative flow networks ~\cite{gflowbeng} have recently been employed to solve CO problems in an unsupervised fashion. DiffUCO ~\cite{diffuco} describes a diffusion-based model, where during the forward process, discrete solutions are iteratively corrupted with noise until they converge towards a stationary distribution. Subsequently, during the reverse process, a GNN is trained to iteratively remove the noise and generate good complete configurations (discrete solutions). On the other hand, a generative flow network (GflowNet) ~\cite{gflownets} designs a problem-specific Markov decision process (MDP) to sample good complete configurations from the distribution of solutions in an autoregressive manner. Both generative approaches were tested on standard benchmark CO problems.

\textbf{CSP via QUBO/HUBO.}~\cite{gusev2023optimality} formulated the CSP task as an integer programming optimization problem. The crystal structure is represented as a set of binary variables which express the existence or non-existence of an atom on each grid point of a discretized unit cell. The total crystal energy, which corresponds to the objective function, is defined as the sum of pairwise energy contributions. These can be efficiently computed using two-body interatomic potentials for each grid point and atom type. The objective function takes the form of a quadratic unconstrained binary optimization (QUBO) problem and can be solved using branch-and-cut methods, which are implemented in state-of-the-art optimizers~\cite{gurobi}. Similarly,~\cite{couzinie:23} proposed an annealing scheme usable on modern Ising machines for CSP by taking into account the general $n$-body atomic interactions, and in particular three-body interactions, which are necessary to simulate covalent bonds. The crystal structure is represented in a similar way to~\cite{gusev2023optimality} and the CSP problem can be defined as a quadratic unconstrained binary optimization (QUBO) or higher-order unconstrained binary optimization (HUBO), where additional penalty terms are incorporated in the Hamiltonian to account for the constraints. A GPU-based Ising machine, Fixstars Amplify, was used in~\cite{Liang2025crysim} for CSP, with a specific focus on large crystals with high symmetry.

\section{Problem Definition} \label{prob_def}
To mathematically formalize the optimization problem of CSP, we follow the notation used by~\cite{gusev2023optimality}. Let $\mathcal{P} = \{p_1, p_2, \ldots, p_n\}$ be a set of $n$ positions within a unit cell. These positions constitute a lattice and are described by their fractional coordinates as $(\frac{i}{g}, \frac{j}{g}, \frac{k}{g})$, with integers $i, j, k \in (0,1, \dots, g-1)$.  
The discretization parameter $g$ defines the number of discrete positions along each dimension of the unit cell. We also let $\mathcal{T} = \{t_1, t_2, \ldots, t_k\}$ be a set of $k$ chemical elements under consideration and the crystal stoichiometry $\mathbf{c} = [c_1, c_2, \ldots, c_k]$, the number of atoms of each respective element. CSP aims to find the allocation of a given number of atoms belonging in $\mathcal{T}$ to positions $\mathcal{P}$ so that the total interaction energy is minimized. As the number of atoms can be lower than the number of positions, some subset of the positions can be left empty (void). An example is given in Figure~\ref{fig:schematic_figure}, where atoms of types $\mathcal{T}=\{\operatorname{Sr}, \operatorname{Ti}, \operatorname{O}\}$ with respective stoichiometries $\mathbf{c}=[1,1,3]$, have to be assigned to a dense grid of 343 ($g=7$) positions within the unit cell.

To measure the energy of an allocation, we use methods based on interatomic potentials \cite{article}. These methods define the energy as the sum of two terms, namely the electrostatic interaction of ions and the repulsion from close ions, with the ions regarded as point charges. Since the electrostatic interaction is long-range, it can be computed using the Ewald summation method \cite{Toukmaji1996EwaldST}. The Ewald sum can be expressed as a finite sum over all atom pairs within a unit cell, where each pairwise contribution can be computed independently of the positions of other atoms. 
By representing the repulsion term using two-body potentials (such as Buckingham or Lennard-Jones), the total potential energy can be formulated as the sum of pairwise contributions. Each of these accounts for the allocation of a pair of ions. Importantly, the contribution of each possible ion pair allocation can be computed independently and stored in a table. Assuming that ions with chemical elements $t_1$ and $t_2$ are assigned to positions $p_1$, $p_2$, respectively, $\alpha_{p_1, p_2}^{t_1, t_2}$ corresponds to their pairwise energy contribution. By introducing binary variables $X^{t}_{p}$ for each chemical element $t$ and position $p$ (with $X^{t}_{p}=1$ if element $t$ is placed at position $p$ and zero otherwise), the total interaction energy can be calculated as \cite{gusev2023optimality}:

\begin{equation}
  E = \sum_{\substack{p_1, p_2 \in \mathcal{P}, \\ t_1, t_2 \in \mathcal{T}}}  \alpha_{p_1, p_2}^{t_1, t_2} X^{t_{1}}_{p_{1}} X^{t_{2}}_{p_{2}}, \quad X \in \{0, 1 \},
  \label{Eq_1}
\end{equation}

Eq.~\ref{Eq_1} consists of quadratic terms and takes the form of the objective function of a quadratic unconstrained binary optimization problem (QUBO)~\cite{qubo}. CSP, as an atom allocation problem, has exclusivity and stoichiometry constraints. Exclusivity ensures that each position is allocated with at most one atom, while stoichiometry guarantees that the number of allocated atoms for each chemical element corresponds to the element's specified stoichiometry value.

One way to account for these constraints is to introduce appropriate penalty terms and fold them into the objective of eq.~\ref{Eq_1}, as done by~\cite{gusev2023optimality}. However, this results in additional computational overhead, as these terms come with respective penalty weights that need to be tuned carefully and often specifically for each composition and due to the need for verifying the feasibility of computed solutions.

To avoid the need for penalty terms, we reformulate eq.~\ref{Eq_1} by defining the allocation matrix $\mathbf{X} \in \mathcal{R}^{n\times k}$ which describes the hard (one-hot-encoded) assignment of atoms from $\mathcal{T}$ with stoichiometry $\mathbf{C}$ to $n$ positions. Each row corresponds to a specific position, while each column corresponds to a specific chemical species. To mathematically encode the void possibility for each position, we introduce an additional column to \textbf{X} which becomes $\mathbf{X} \in \mathcal{R}^{n\times (k+1)}$ and comes with a stoichiometry value of $n-\sum_{i=1}^{k} c_i$ that corresponds to the number of empty positions and is included as the $(k+1)$-th value of the updated stoichiometry vector $\mathbf{c}$. The exclusivity constraint is satisfied by ensuring that the rows of \textbf{X} sum to 1, while the correct stoichiometry is satisfied by ensuring that the columns sum to the correct atom-specific stoichiometry given by $\mathbf{c}=[c_1, c_2, \ldots, c_k,n-\sum_{i=1}^{k} c_i]$. Finally,  $\operatorname{vec}(\mathbf{X})$ corresponds to its column-vectorized version, and we have:
\begin{equation}
\begin{aligned}
  E(\mathbf{X}) &= \operatorname{vec}(\mathbf{X})^{\top}\mathbf{Q}\,\operatorname{vec}(\mathbf{X}), \\
  \text{s.t.}\quad & \mathbf{X} \in \{0, 1\}^{n\times (k+1)}, \; 
  \mathbf{X}^{\top}\mathbf{1}_{n} = \mathbf{1}_{k+1}, \; 
  \mathbf{X}\mathbf{1}_{k+1} = \mathbf{c}
\end{aligned}
\label{Eq_2}
\end{equation}

Here $\mathbf{1}_n$ corresponds to a column vector of length $n$ whose elements are equal to 1. $\mathbf{Q} \in \mathcal{R}^{n(k+1) \times n(k+1)}$ stores all possible pairwise energy contributions for each position and chemical element. We wish to compute atom allocations $\mathbf{X}$ that simultaneously minimize the objective of eq.~\ref{Eq_2} and satisfy both the exclusivity and stoichiometry constraints.

\section{Preliminaries} \label{preliminaries}
In this section, we briefly introduce few concepts that are integral parts of our proposed approach and contributions. More specifically, we later introduce a graph construction that is inspired by expander graphs. Furthermore, we train a graph neural network in an end-to-end fashion to compute node representations that are projected in the transportation polytope of feasible CSP solutions using an appropriately designed Sinkhorn operator.

\subsection{Matrix scaling problem and transportation polytopes} \label{polytopes}
The matrix scaling problem looks to scale a non-negative matrix so that it has fixed positive prescribed row and column sums. It can be solved using the Sinkhorn-Knopp algorithm~\cite{83900530-3d01-3dd8-a11c-d939e82fce35}, which is applied to a matrix by iteratively alternating between scaling its rows and its columns. The formal definition of a matrix scaling problem and the Sinkhorn-Knopp algorithm are provided in section~\ref{matrix scaling-sinkhorn} of the Appendix. Importantly, the resulting scaled matrix belongs to a transportation polytope.

Formally, a transportation polytope consists of non-negative $\mathbf{P}  \in \mathbb{R}_{\ge 0}^{n \times m}$ matrices, termed as feasible transport plans, that have fixed positive row sums $\mathbf{r} \in \mathbb{R}^n>0$ and column sums $\mathbf{c} \in \mathbb{R}^m>0$, and can be written as:

\[
\mathcal{U}(\mathbf{r},\mathbf{c})
=
\bigl\{
\mathbf{P} \in \mathbb{R}_{\ge 0}^{n \times m}
\;\big|\;
\mathbf{P} \mathbf{1}_m = \mathbf{r},\;
\mathbf{P}^\top \mathbf{1}_n = \mathbf{c}
\bigr\}.
\]
In the case of $\mathbf{r} \in \mathbb{Z}_{>0}^n, \mathbf{c} \in \mathbb{Z}_{> 0}^m$, the feasible transport plan matrices are integral, i.e. $\mathbf{P} \in \mathbb{Z}_{\geq0}^{n \times m}$, also denoted as basic feasible solutions (BFS) and correspond to the vertices of $\mathcal{U}(\mathbf{r},\mathbf{c})$.

\subsection{Geometric graph construction and expander graphs}
In the CSP context, the crystal data define an atomic system that consists of a set of atom types (as defined in the stoichiometry constraint) and a set of 3D positions, which form a 3D point cloud and serve as candidates for the assignment of the atoms. Crystals can be modeled by infinite periodic structures, with the unit cell being the smallest repeating three-dimensional substructure. A geometric crystal graph represents the unit cell by considering neighboring cells in all directions using periodic boundary conditions (PBC). These appropriately modify the calculation of the nodes' distances to account for neighboring cells. To construct a geometric graph from a unit cell, we take its points as nodes, and we need to define an adjacency matrix that models pairwise interactions while considering the PBC.

Popular approaches for geometric graph construction involve complete graphs, cutoff graphs, and long-range connections ~\cite{duval2024hitchhikersguidegeometricgnns}. In complete graphs, every node is connected to every other node, and thus all pairwise interactions (long-range among them) can be captured. Although this strategy has been chosen for small molecular datasets (MD17~\cite{10.5555/2969442.2969488}), it comes with increased computational complexity as the graphs become very dense. This issue is addressed in radius cutoff graphs ~\cite{10.5555/3294771.3294866}, ~\cite{10.5555/3540261.3540781}, where nodes are connected with an edge if their Euclidean distance is below a cutoff threshold distance, often measured in Angstrom units. To make a radius cutoff graph regular, a fixed number of nearest neighbors is selected for each node. Due to the local nature of cutoff graphs, GNNs often struggle to aggregate messages among distant nodes without significant loss of information. Although stacking more layers helps to consider long-range connections that can model electrostatic interactions and van der Waals forces, this often leads to over-smoothing. Relevant approaches account for this shortcoming of cutoff graphs by enriching them with long-range connections that could be randomly sampled with some heuristically chosen probability ~\cite{ingraham}.

Meanwhile, even after dealing with over-smoothing, GNNs can still struggle with incorporating global information from distant nodes, as a result of over-squashing~\cite{oversquashing}. Informally, over-squashing can be deemed as the difficulty in exchanging messages between two nodes
without loss of information and can happen when the reachability of two nodes is limited (there are only few possible routes between them in contrast to every other available route for the rest of nodes). Intuitively, over-squashing occurs when bottlenecks appear in a graph.
Bottlenecks correspond to a small number of edges that are under representational pressure~\cite{deac2022expandergraphpropagation} because they have to transport information between large groups of nodes. An example of a bottleneck is shown in figure~\ref{fig:barbell} in the Appendix.

Expanders~\cite{hoory06} are a special kind of graphs with desirable properties, such as a small diameter, that are sparse, having edges linear in the number of nodes, but at the same time highly interconnected. Intuitively, in an expander graph, any two vertices are connected by a path consisting of a few hops. Importantly, expanders allow effective communication between nodes, eliminating topological properties known as bottlenecks that are often responsible
for over-squashing and providing a paradigm that enables global propagation of information with message-passing neural networks. In addition to that, their sparsity helps reduce the complexity and computational demands.

\textbf{Margulis graph.}
~\cite{Margulis1973} introduced the first explicit expander graph construction. For an integer $n$, we let $\mathbb{G}_n$ be a graph with $n^2$ vertices and vertex set $\mathbb{Z}_n \times \mathbb{Z}_n$. Then, we can construct the Margulis expander by selecting as neighbors of each vertex $(x, y) \in \mathbb{Z}_n \times \mathbb{Z}_n$, the following vertices: 
\begin{equation}
(x\pm 1, y ), (x, y \pm 1), (x\pm y, y), (x, y\pm x),
\label{eq:margulis}
\end{equation}
where all operations are $\operatorname{modulo} n$. As a result, $\mathbb{G}_n$ is an 8-regular graph with parallel edges and self-loops.

\textbf{Gabber-Galil graph.}
Subsequently,~\cite{gabber} provided an analysis of~\cite{Margulis1973} graphs and introduced an additional graph construction with expansion properties better than the Margulis one. This construction, following the same notation as the margulis one, selects as neighbors of vertex $(x,y,z)$ as :

\begin{equation}
(x, y \pm 2x), (x, y \pm (2x+1)), (x\pm 2y, y), (x\pm (2y+1), y)
\label{eq:gabber-galil}
\end{equation}
As in the Margulis case, all operations are $\operatorname{modulo} n$.

\subsection{Graph Neural Networks}
Graph neural networks are deep learning models that are applied to graph data. GNNs compute node representations under permutation invariance (with respect to node ordering) by implementing the message-passing process for a fixed number of iterations that correspond to the number of GNN layers. This process updates the representation of each node by aggregating feature information from its local neighbors.

More formally, let $\mathcal{G}=(\mathcal{V}, \mathcal{E})$ be a graph with node features $\mathbf{X} \in \mathbb{R}^{d \times n}$. Then, the update rule for the $k^{th}$ layer of the GNN is defined as \cite{messagepassing}:
\begin{equation}
\begin{aligned}
\mathbf{m}_v^k &= \operatorname{AGGREGATE}_w^k(\{\mathbf{h}_u^{k-1} \mid u \in \mathcal{N}_v\}), \\
\mathbf{h}_v^k &= \operatorname{UPDATE}_w^k(\mathbf{h}_v^{k-1}, \mathbf{m}_v^k)
\end{aligned}
\label{eq:message-passing}
\end{equation}
where $\mathbf{h}_v^k$ corresponds to the representation (embedding) of node $v$ computed at layer $k$, with $\mathbf{h}_v^0=\mathbf{x}_v$ being the $d$-dimensional initial feature vector of the node. Furthermore, $\operatorname{AGGREGATE_w(\cdot)}$ and $\operatorname{UPDATE_w(\cdot)}$ are parameterizable differentiable functions ~\cite{Kipf:2016tc}, ~\cite{velikovi2017graph}, ~\cite{xu2018how}. For each layer $k$, a vectorized ``message" $\mathbf{m}_v^k$ is computed for every node $v$ by aggregating information from its local neighborhood defined as $\mathcal{N}_v=\{u \in \mathcal{V}|(u,v) \in \mathcal{E} \}$. Intuitively, by stacking $K$ layers, every node embedding considers information from its $K$-hop neighborhood, also denoted as the receptive field of the graph. The resulting node embeddings $\mathbf{h}_v^K$ can be directly fed into a task-specific loss function and, along with the network's parameters, can be optimized in an end-to-end fashion (usually with some form of stochastic gradient descent) to solve tasks such as classification and prediction.

\section{Methods}
In this section, we present our two main contributions, which include a differentiable method to solve the CSP problem and the construction of an expander-inspired graph from a 3D point cloud of positions that represent the discretized unit cell. More specifically, we introduce an approach which inherently handles both CSP constraints, as a result of which eliminating the need for penalty terms to account for them. To do this, we define an appropriate Sinkhorn operator and combine it with differentiable sampling techniques to sample and obtain feasible optimal structures. Furthermore, we introduce an expander-inspired 3D graph which consists of both local and long-range interactions and provides a better choice for message-passing than radius cutoff graphs.

\subsection{Graph neural transportation for crystal structure prediction}
Following the CSP formulation in eq.~\ref{Eq_2} and the definition of transportation polytopes in section~\ref{polytopes}, the CSP solutions belong to the vertices of the transportation polytope $\mathcal{U}(\mathbf{r},\mathbf{c})$. This polytope consists of matrices with row sums $\mathbf{r}=[1, \ldots,1]=\mathbf{1}_n$ and column sums $\mathbf{c}=[c_1, c_2, \ldots, c_k,n-\sum_{i=1}^{k} c_i]$, encoding the exclusivity and stoichiometry constraints in the CSP context, respectively. We wish to compute the optimal basic feasible solution that minimizes the energy of eq.~\ref{Eq_2}. To do this, we construct a graph (either a radius cutoff one or our expander-inspired) from the 3D point cloud of the discretized unit cell positions and solve CSP as a multiclass node classification problem using a differentiable unsupervised training process that projects the embeddings computed by a GNN to the vertices of $\mathcal{U}(\mathbf{r},\mathbf{c})$.

We start with some randomly initialized node embeddings $h_v^0$ that can be created through an embedding block and are learnable. We then apply a 3-layer graph isomorphism network (GIN)~\cite{xu2018how}, which we modified to incorporate edge features as~\cite{hu2020pretraining}, to compute node embeddings $h_v^l$ that reflect their $l$-hop neighborhood information. At the core, the node embeddings of each layer are calculated after aggregating feature information from their neighbors and applying a nonlinear activation function. Formally, the $l$-th layer of our GNN is:

\begin{equation}
\mathbf{h}_v^{(l)} = \operatorname{GeLU}\!\left( \operatorname{MLP}\! ^ {(l)} \left( (1 + \alpha) \cdot \mathbf{h}_v^{(l-1)} + \sum_{j \in \mathcal{N}(v)} \mathrm{GeLU} (\mathbf{h}_j^{(l-1)} + \mathbf{e}_{j,v}^{(l-1)} ) \right) \right)
\label{eq:gine_update}
\end{equation}
where $\alpha$ is a learnable parameter and $\operatorname{MLP}$ denotes a Multilayer Perceptron. $\mathbf{h}_v^{(0)}$ are randomly initialized via an embedding block. Edge features correspond to vectors that include every possible pairwise energy contribution as a result of placing every combination of atoms in the respective positions, as defined through $\mathbf{Q}$. These features are then passed through a linear layer to match the dimensions of the node embeddings. A schematic diagram of the developed architecture is shown in Figure~\ref{fig:gnn_model} of the Appendix. 

We apply a relaxation strategy to the interaction energy of the CSP problem, as formalized in eq.~\ref{Eq_2}, and replace the (hard) one-hot-encoded assignment matrix with a corresponding (soft) normalized assignment matrix $\mathbf{S} \in [0, 1]^{n\times (k+1)}$. The differentiable CSP problem is formalized as:

\begin{equation}
\begin{aligned}
  E(\mathbf{S}) &= \operatorname{vec}(\mathbf{S})^{\top}\mathbf{Q}\,\operatorname{vec}(\mathbf{S}), \\
  \text{s.t.}\quad & \mathbf{S} \in [0, 1]^{n\times (k+1)}, \; 
  \mathbf{S}^{\top}\mathbf{1}_{n} = \mathbf{1}_{k+1}, \; 
  \mathbf{S}\mathbf{1}_{k+1} = \mathbf{c}
\end{aligned}
\label{Eq_4}
\end{equation}

\textbf{S} needs to provide an approximation of the discrete assignment matrix. A temperature-dependent softmax function, defined as $\operatorname{softmax_{\tau}}(\mathbf{s})_i=\operatorname{exp(\mathbf{s}_i/\tau)}/\sum_j\operatorname{exp(\mathbf{s}_j/\tau)}$, is a normalization operator that can be used to approximate a discrete category from a set of continuous values~\cite{maddison2017the},~\cite{jang2017categorical}. As $\tau \rightarrow 0$, in the limit $\operatorname{softmax_{\tau}}$ converges to a one-hot vector corresponding to the largest $\mathbf{s}_i$. In this section, we introduce a temperature-dependent Sinkhorn operator, which provides an analog of $\operatorname{softmax}$ normalization~\cite{mena2018learning}, and can be used to approximate the choice of a (discrete) basic feasible solution to our CSP problem.

To do this, we first consider~\cite{Sinkhorn1967ConcerningNM} and introduce the Sinkhorn operator $S(\mathbf{\cdot})$ for differentiable CSP, which we apply on the embeddings' matrix $\mathbf{H} \in \mathbb{R}^{n \times (k+1)}$ computed by the GNN. This operator solves a matrix scaling problem by iteratively scaling the rows and columns of $\mathbf{H}$ so that it has the prescribed row $\mathbf{r}$ and columns $\mathbf{c}$ sums, and is defined as : 
\begin{equation}
\begin{split}
 S^0(\mathbf{H}) &= \operatorname{exp}(\mathbf{H}), \\
 S^l(\mathbf{H}) &= \mathcal{T}_c(\mathcal{T}_r(S^{l-1}(\mathbf{H}))),\\
 S(\mathbf{H})   &= \lim_{l\to\infty} S^l(\mathbf{H})
\end{split}
\label{Eq_3}
\end{equation} 
with $T_r(\mathbf{H})
= \mathbf{H} \odot \Big( \big( \mathbf{1}_n \oslash (\mathbf{H}\mathbf{1}_{k+1}) \big)\, \mathbf{1}_{k+1}^{\top} \Big)$, $\mathcal{T}_c(\mathbf{H})=\mathbf{H} \odot \Big(  \mathbf{1}_n \big(\mathbf{c} \oslash (\mathbf{1}_n^{\top} \mathbf{H}) \big)^{\top} \Big)$ denoting row-wise and column-wise scaling operators, respectively, while $\odot$ corresponds to Hadamard product and $\oslash$ to element-wise division. Importantly, following~\cite{adams2011rankingsinkhornpropagation}, we can apply an incomplete version of the Sinkhorn operator, which instead of $l$ has a fixed number of iterations $L$. For numerical stability, the Sinkhorn operator is often implemented in the log-space and exponentiation takes place after convergence. ~\cite{83900530-3d01-3dd8-a11c-d939e82fce35} proved that $S(\mathbf{H})$ belongs to the transportation polytope $\mathcal{U}(\mathbf{r},\mathbf{c})$. Nevertheless, as mentioned above, we wish to compute basic feasible solutions that lie on the vertices of this polytope.

We can select a category from a set of continuous variables $\mathbf{h}$ by finding the one hot vector $\mathbf{v}^*$ that maximizes $\langle \mathbf{h} , \mathbf{v} \rangle$, i.e. the index of the largest value of $\mathbf{h}$. In a similar manner, we parameterize the hard (discrete) choice of a basic feasible solution $\mathbf{B}$ through a matrix $\mathbf{H}$ as the solution to the linear assignment problem~\cite{kuhn}, with $\operatorname{vert(\mathcal{U}(\mathbf{r},\mathbf{c}))}$ denoting the set of vertices of the transportation polytope $\mathcal{U}(\mathbf{r},\mathbf{c})$ and $\langle \mathbf{A} , \mathbf{B} \rangle_F=\operatorname{trace(\mathbf{A},\mathbf{B}^T)}$ denoting the Frobenius inner product of matrices, as:
\begin{align}
\label{eq:lap_tp}
 \operatorname{BFS}(\mathbf{H}) =  & \argmax_{\mathbf{B}\in \operatorname{vert(\mathcal{U}(\mathbf{r},\mathbf{c}))}} \langle \mathbf{B} , \mathbf{H} \rangle_F
\end{align}

~\cite{cuturi} used entropic regularization to prove that, in the limit, for a small $\tau$, $S(\mathbf{H} / \tau)$ approximates $\operatorname{BFS}(\mathbf{H})$, i.e., $\operatorname{BFS}(\mathbf{H}) \approx S(\mathbf{H} / \tau)$ and can thus serve as a differentiable relaxation of it. This is summarized in the theorem provided in section~\ref{entropic} of the Appendix. In practice, choosing lower $\tau$ values usually requires more Sinkhorn iterations to achieve convergence as the solution is encouraged to converge to a basic feasible solution, which is in essence one-hot encoded. As a result of this theorem, we can use our Sinkhorn operator to compute basic feasible solutions in a differentiable fashion.

Furthermore, techniques similar to the re-parameterization trick~\cite{Kingma2014} can be applied for the differentiable sampling from latent distributions. Importantly, discrete distributions can become re-parameterizable via the gumbel trick~\cite{10.1109/ICCV.2011.6126242}, by taking $\operatorname{argmax}(\mathbf{h} + \epsilon)$, where $\epsilon \sim \operatorname{Gumbel}$. To make this operation differentiable, we replace $\operatorname{argmax}$ with $S(\cdot)$. Overall, to parameterize the sampling from the latent distribution of basic feasible solutions, we introduce i.i.d. samples of $\operatorname{Gumbel}$ noise $\epsilon$ to the entries of $\mathbf{H}$ and apply our Sinkhorn operator for differentiable CSP as $S(\mathbf{H + \epsilon}) / \tau$. As a result, we can differentiably sample from the distribution of feasible atom allocations that satisfy both CSP constraints. After Sinkhorn convergence, the resulting matrix is exponentiated in an entry-wise fashion and corresponds to the soft assignment matrix $\mathbf{S}$ of a CSP atom allocation. Finally, we apply a simple threshold of $0.5$ to the soft assignment probabilities $\mathbf{S}$ and map them to the one-hot-encoded allocation matrix $\mathbf{\Sigma}$, i.e., $\mathbf{\Sigma}_{ij}=[\mathbf{S}_{ij} \geq 0.5]$. The complete pipeline of GNT-CSP is provided in algorithm~\ref{alg:differentiable-csp}. 

To conclude, using our end-to-end differentiable approach, we can compute atom allocations that satisfy both constraints of CSP and look for  optimal feasible structures that minimize the total interaction energy.

\begin{algorithm}[H]
\caption{GNT-CSP}
\label{alg:differentiable-csp}
\begin{algorithmic}[1]
\REQUIRE Stoichiometry constraint vector $\mathbf{c}$, Pairwise interactions matrix $\mathbf{Q}$, temperature value $\tau$,  number of epochs \texttt{Epochs}, number of positions $n$, number of unique atom species $k$, maximum number of Sinkhorn iterations \texttt{SinkIter}, sinkhorn convergence precision \texttt{d}, number of GNN layers \texttt{L}

\ENSURE Energy of the best feasible solution to CSP $E_{\mathrm{best}}$

\STATE Construct graph $\mathcal{G}$ with $n$ nodes 
\STATE Initialize node embeddings $\mathbf{h}_v^{(0)}$ via an embedding block
\STATE Initialize edge embeddings $\mathbf{e}^{(0)}$ using $\mathbf{Q}$
\STATE $E_{\mathrm{best}} \gets \infty$
\FOR{$e = 1$ to \texttt{Epochs}}
    \FOR{$l=1$ to \texttt{L}}
   \STATE $\mathbf{h}_v^{(l)} = \operatorname{GeLU}\!\left( \operatorname{MLP}\! ^ {(l)} \left( (1 + \alpha) \cdot \mathbf{h}_v^{(l-1)} + \sum_{j \in \mathcal{N}(v)} \mathrm{GeLU} (\mathbf{h}_j^{(l-1)} + \mathbf{e}_{j,v}^{(l-1)} ) \right) \right)$ \COMMENT{Compute node embeddings with message-passing on $\mathcal{G}$}
    \ENDFOR
\STATE $\mathbf{\epsilon} \sim \mathrm{Gumbel}(0,1)$
\STATE $\mathbf{H} = (\mathbf{H}^L + \mathbf{\epsilon}) / \tau$ 
    \FOR{$i=1$ to \texttt{SinkIter}}
    \STATE $\mathbf{H} = \mathbf{H} - \log(\sum_{j=1}^{k+1} \exp{\mathbf{H}_{j}})$ \COMMENT{Vectorized row-scaling}
    \STATE $\mathbf{H} = \mathbf{H} + \log \mathbf{c} - \log(\sum_{i=1}^{n} \exp{\mathbf{H}_{i}})$ \COMMENT{Vectorized column-scaling}
    \IF{$|| \operatorname{exp} (\mathbf{H}) \textbf{1}_{k+1}-\mathbf{1}_n) ||_{\operatorname {\infty}} < d$ \text{and} $|| \operatorname{exp}(\mathbf{H})^ \top \textbf{1}_n - \mathbf{c} ||_{\operatorname {\infty}} < d$}
    \STATE $\mathbf{S} = \operatorname{exp}(\mathbf{H})$ \COMMENT{Exponentiation after convergence to get soft allocation matrix} 
    \STATE break \COMMENT{Sinkhorn termination}
    \ENDIF
    \ENDFOR
    \STATE $E(\mathbf{S}) = \operatorname{vec}(\mathbf{S})^\top \mathbf{Q} \operatorname{vec} (\mathbf{S})$ \COMMENT{Calculate soft atom allocation energy}
    \STATE $\operatorname{min}E(\mathbf{S})$ \COMMENT{Update GNN's weights using backpropagation}
    \STATE $\mathbf{\Sigma} = \mathbf{S} \geq 0.5$ \COMMENT{Compute one-hot encoded allocation matrix}
    \STATE $E(\mathbf{\Sigma}) = \operatorname{vec}(\mathbf{\Sigma})^\top \mathbf{Q} \operatorname{vec} (\mathbf{\Sigma})$ \COMMENT{Calculate atom allocation energy}
    \IF{$E(\mathbf{\Sigma})<E_{\mathrm{best}}$ and $\mathbf{\Sigma} \mathbf{1}_{k+1} = \mathbf{1}_n$ and $\mathbf{\Sigma} ^ \top \mathbf{1}_n = \mathbf{c}$}
        \STATE $E_{\mathrm{best}}=E(\mathbf{\Sigma})$ \COMMENT{Update best feasible solution energy}
    \ENDIF
\ENDFOR
\RETURN $E_{\mathrm{best}}$
\end{algorithmic}
\end{algorithm}

\subsection{Geometric graph construction}
Expander graphs~\cite{hoory06} are sparse and provide a spectral approximation of a complete graph~\cite{shirzad2023exphormer}. They are designed to have strong connectivity properties (such as low diameter and good mixing)~\cite{deac2022expandergraphpropagation}, allowing efficient information propagation between distant nodes within a small number of hops, which helps eliminate bottlenecks and can alleviate the over-squashing phenomenon of GNNs. Their sparsity, which ensures a lower complexity and thus lighter computational requirements, and their desirable connectivity properties motivate the introduction of a 3D graph inspired by expanders that can be used for message-passing instead of local cutoff graphs.

\textbf{Proposed graph construction}. Our proposed graph construction extends the connectivity pattern of the Gabber-Galil graph~\cite{gabber} to 3D. Formally, for an integer $n$, we let $\mathcal{G}_n$ be a graph that has $n^3$ vertices with vertex set $\mathbb{Z}_n \times \mathbb{Z}_n \times \mathbb{Z}_n$. We select the neighbors of each vertex $(x,y,z) \in  \mathbb{Z}_n \times \mathbb{Z}_n \times \mathbb{Z}_n$, by keeping one at a time coordinate fixed and using the Gabber-Galil formulas of Eq.~\ref{eq:gabber-galil} on the other two. We repeat this procedure for all three pairs of coordinates. In other words, we apply the Gabber-Galil construction on the 'xy', 'yz' and 'xz' planes separately, and get the following neighbors for each vertex $(x,y,z)$ :

\begin{equation}
\begin{aligned}
(x, y \pm 2x, z), (x, y \pm (2x+1), z), (x\pm 2y, y, z), (x\pm (2y+1), y, z) \\
 (x, y \pm 2z, z), (x, y \pm (2z+1), z), (x, y, z\pm 2y), (x, y, z\pm (2y+1))\\
 (x\pm 2z, y , z), (x \pm (2z+1), y, z), (x, y, z\pm 2x), (x, y, z\pm (2x+1)) 
\end{aligned}
\label{eq:3d-gabber-galil}
\end{equation}
Finally, we make sure that there are no parallel edges, which means that each edge appears only once. The resulting graph is sparse and designed to provide a good balance between local and longer-range connections.

\section{Evaluation}

\subsection{Experimental setup}
Our experiments involve solving the combinatorial CSP on a set of compositions, namely \ch{SrTiO3}, \ch{Y2O3}, \ch{Y2Ti2O7}, \ch{MgAl2O4} and  \ch{Ca3Al2Si3O12}, which correspond to cubic crystal structures of perovskite, bixbyite, pyrochlore, spinel and garnet types, respectively. These compositions were modeled using classical pairwise force-fields, with their parameters provided in~\cite{gusev2023optimality}. Importantly, unlike~\cite{gusev2023optimality}, we solve the periodic lattice atom allocation problem, using the simplest symmetry group P1(1) containing only identity operation. Therefore, all $g^3$ lattice positions are unique, where $g$ defines the number of lattice positions per side of the unit cell. We investigate different supercells of \ch{SrTiO3} by varying the number of formula units (Z), which controls how much the unit cell is expanded in each dimension. Increasing Z leads to additional symmetries and more atoms within the unit cell, which increases the CSP complexity.

To measure the difficulty in solving the CSP of different compositions, we calculate the size of the combinatorial space, which refers to the size of the set that includes all feasible configurations of a composition. This can be computed by the multinomial coefficient $\binom{n}{c_1,c_2, ... , c_k}=\frac{n!}{c_1!c_2!...c_k!}$, where $c_i$ correspond to the stoichiometry constraints, as defined in Section~\ref{prob_def}. Table~\ref{tab:compound_properties} reports multiple properties for the examined compositions as well as the order of magnitude of the combinatorial space for the periodic atom allocation of each composition.

\begin{table}[ht]
\centering
\renewcommand{\arraystretch}{1.2} 
\setlength{\tabcolsep}{5pt} 
\begin{tabular}{|l|c|c|c|c|c|}
\hline
\textbf{Compound} & \makecell{\textbf{Number of ions} \\ \textbf{in the unit cell}} & \textbf{Cell parameter (\AA)} & \textbf{Discretization, g} & \makecell{\textbf{Number of unique} \\ \textbf{lattice positions}} & \makecell{\textbf{Combinatorial} \\  \textbf{Space Size}} \\ \hline
\ch{SrTiO3}, Z=1 & 5 & 3.9 & 4 & 64 & $10^8$ \\ \hline
\ch{SrTiO3}, Z=1 & 5 & 3.9 & 8 & 512 & $10^{12}$ \\  \hline
\ch{SrTiO3}, Z=8 & 40 & 7.8 & 8 & 512 & $10^{74}$ \\ \hline
\ch{SrTiO3}, Z=27 & 135 & 11.7 & 8 & 512 & $10^{180}$ \\ \hline
\ch{Y2O3} & 80 & 10.7 & 8 & 512 & $10^{97}$ \\ \hline
\ch{Y2Ti2O7} & 88 & 10.2 & 8 & 512 & $10^{133}$ \\ \hline
\ch{MgAl2O4} & 56 & 8.2 & 8 & 512 & $10^{97}$ \\ \hline
\ch{Ca3Al2Si3O12} & 160 & 11.9 & 8 & 512 & $10^{210}$ \\ \hline
\end{tabular}
\caption{The list of compositons used in our computational experiments and the parameters of atom allocation problems. }
\label{tab:compound_properties}
\end{table}


We compare the performance of multiple algorithms in solving the CSP problem across all reported compositions. The algorithms used include our proposed approach, IPCSP (~\cite{gusev2023optimality}), Simulated Annealing (SA), and a bespoke Greedy solver.
We evaluate the quality of each obtained structure by examining the energy associated with it. Furthermore, since the ground truth energy of each composition has been reported in ~\cite{gusev2023optimality}, we use the relative optimality gap (ROG) to measure how close a configuration with energy $E_f$ is to the ground state with energy $E_g$, expressed as a percentage of the ground state energy. Mathematically, this is defined as:
\begin{equation}
    \operatorname{ROG}=\frac{|E_g-E_f|}{|E_g|},
    \label{eq:rog}
\end{equation}
where lower ROG values are indicative of higher-quality allocations whose energies are closer to the ground state. Note that, for some discretizations, an ROG of zero is unattainable.

\textbf{GNT-CSP}. We implemented our method and ran experiments with it on the proposed graph. To highlight the gain of introducing the expander-inspired graph, we additionally ran experiments using radius cutoff graphs (using an empirically selected radius cutoff threshold of 4 \AA~and 16 nearest neighbors) and an extension of the Margulis graph in 3D for message-passing and provide their results in Section~\ref{radius_margulis} of the Appendix. Our method has been implemented in Python, using the open-source libraries PyTorch and PyTorch geometric ~\cite{Fey/Lenssen/2019} and is publicly available in \url{https://github.com/StavGer/CSP-with-GNNs}. To tune the GNN hyperparameters, we performed five searches (each on three seeds), using hyperopt ~\cite{hyperopt} on the CSP of \ch{Ca3Al2Si3O12}, whose structure is the most difficult to predict with respect to the combinatorial search space. Finally, for each composition, we trained our model five times, for 100k epochs each, and report the lowest energy and ROG of all computed structures. To run our experiments, we used an A100 GPU with 80 GiB of memory.

\textbf{IPCSP}. We used the official IPCSP implementation, which solves the CSP task via Gurobi, provided by ~\cite{gusev2023optimality} and set P1(1) as the symmetry group for each structure, as well as a 24 hour time limit for the termination of each run.


\textbf{Greedy}. As a simple baseline we implemented a greedy method to solve the CSP task in its unconstrained form, where we start from a random feasible configuration that satisfies both the exclusivity and the stoichiometry constraints. We then sample a new configuration that we make sure is consistent with both constraints by moving an atom from an occupied position to an empty one, where both positions are chosen randomly. Greedy adopts a new configuration $\mathbf{x}'$ with energy $E(\mathbf{x}')$ only when it is lower than the energy $E(\mathbf{x})$ of the current configuration $\mathbf{x}$. This process is repeated for a finite number of steps and the configuration with the lowest energy is returned. A greedy pseudoalgorithm for CSP is presented in Algorithm~\ref{alg:greedy-csp} in section~\ref{app:greedy_SA} of the Appendix.


\textbf{Simulated Annealing (SA)}. For Simulated annealing, we sample new configurations in the same fashion as greedy. However, while SA accepts new configurations that have lower energy than the current one, it is additionally designed to accept higher energy configurations with probability based on the temperature-dependent Metropolis criterion~\cite{1953JChPh..21.1087M}. Specifically, assuming that a configuration $\mathbf{x}$ has energy $E(\mathbf{x})$ and a candidate configuration $\mathbf{x}'$ has energy $E(x')$, the probability of accepting the new configuration is defined as :

\begin{equation*}
P(\mathbf{x}'|\mathbf{x}) =
\begin{cases}
1, & \text{if } E(\mathbf{x}') < E(\mathbf{x}), \\[8pt]
\exp\!\left( -\dfrac{E(\mathbf{x}') - E(\mathbf{x})}{T} \right), & \text{if } E(\mathbf{x}') \ge E(\mathbf{x}),
\end{cases}
\label{eq:metropolis}
\end{equation*}

where $T>0$ is the temperature value. Higher temperature values result in accepting most moves to configurations with worse energies, while lower temperature values lead to accepting moves to worse configurations more rarely.

We implement SA and iteratively sample new configurations, which we accept or reject based on the Metropolis criterion. Importantly, during this iterative process, we anneal the temperature from higher to lower values. As a result, SA favors the exploration of the configuration space in the earlier stages of the solution search and allows for escaping local minima, while it switches to exploitation and becomes greedy once the space has been explored to a good extent. Finally, SA comes with a few hyperparameters that need to be appropriately selected. We provide details about their selection as well as context about additional strategies to find new configurations that we experimented with in section~\ref{app:greedy_SA} of the Appendix, where we also include the pseudoalgorithm~\ref{alg:sa-csp} of our SA implementation.

For each composition, we ran greedy and SA ten times (shots), for 200k steps each, and report the lowest energy and ROG of all predicted structures in Table~\ref{tab:main_results}.  

\subsection{Results and Discussion}
Our numerical results are summarized in Table~\ref{tab:main_results}, with the compositions listed in ascending order according to the size of their combinatorial search space. We include the energies of the reference structures reported in~\cite{gusev2023optimality} in the ``Ground Truth'' column. Note that, depending on the discretization, this value can be strictly lower than any feasible solution of the allocation problem. For each composition, we used every method to obtain feasible structures and report their lowest ROG and energy (in eV) associated with the best feasible structure. We observe that all methods find the optimal structure of \ch{SrTiO3} with $Z=1$ for both $g$ values as $\operatorname{ROG}=0$. With the remaining compositions, SA finds better allocations than greedy by 1-6 \%, while IPCSP consistently outperforms both of them, by 1-14 \%. For \ch{MgAl2O4} and \ch{Ca3Al2Si3O12}, our method outperforms the state-of-the-art Gurobi solver used in IPCSP. In addition to this, \ch{SrTiO3} with $Z=8$, which involves more atoms than the $Z=1$ case, can only be optimally predicted by our approach with the Gabber-Galil 3D graph. For the remaining compositions, the allocations computed by IPCSP and our method yield comparable energies. 

\begin{figure}[ht]
    \centering
    \includegraphics[width=0.6\linewidth]{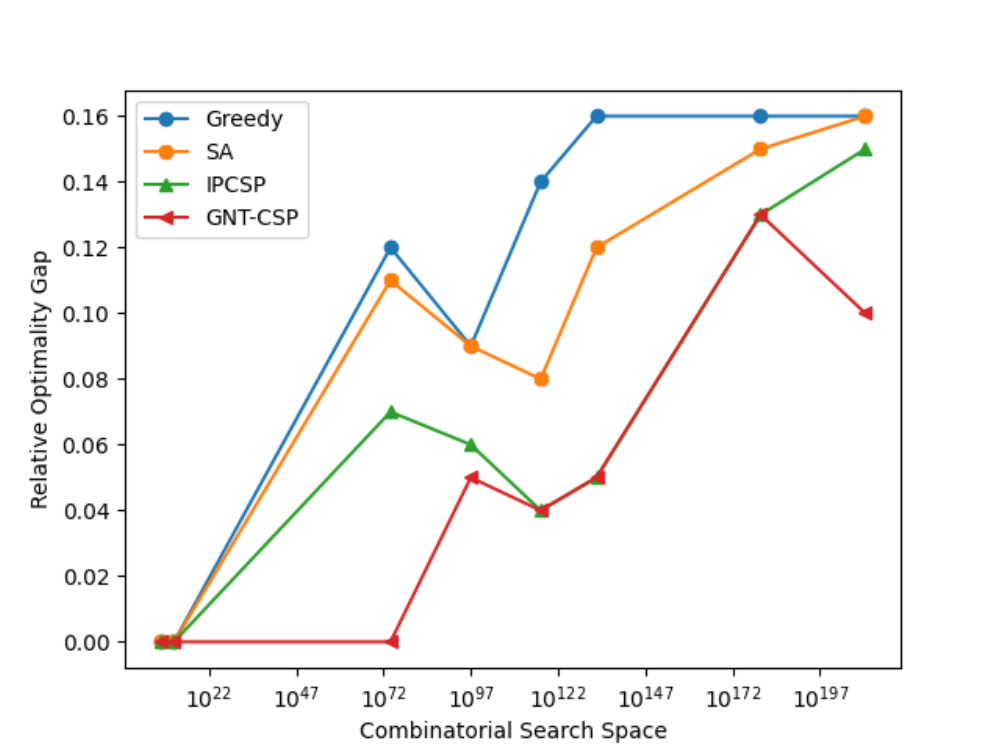}
    \caption{\textbf{ROG curves for the configurations predicted by each method as a function of the combinatorial search space (in logarithmic scale) of the investigated compositions.} The orange curve corresponding to SA is always below the blue curve of greedy
    but above the green curve of the traditional solver used in IPCSP. Therefore, IPCSP performs consistently better than both classical methods SA, while SA outperforms greedy. GNT-CSP performs at least as good as the traditional solver of IPCSP and often better than it for compositions with different magnitudes of combinatorial search spaces, as the red curve is below or intersecting with the green one.}
    \label{fig:rog_main}
\end{figure}

These findings are also illustrated in Figure~\ref{fig:rog_main}, where ROG is plotted as a function of the combinatorial search space for each composition and method. We generally observe that the performance of all methods tends to drop (ROG becomes larger) as the combinatorial search space of the compositions we investigate increases. This is expected because compositions with larger sizes of combinatorial search space include more feasible structures that correspond to local minima and are often difficult for computational methods to escape from. An interesting discussion point concerns the prediction of \ch{SrTiO3} with $Z=8$ which exhibits a strongly repetitive structure inherited from $Z=1$ case and a combinatorial search space with a size magnitude of $10^{72}$. As mentioned above, GNT-CSP is the only one that can compute an optimal structure for it, with all three competitors finding suboptimal structures. In addition to this, all three competitors achieve slightly better performance (lower ROG) when dealing with compositions that have a slightly larger combinatorial search space than $10^{72}$. Although this may seem counterintuitive, it occurs due to this additional structural periodicity present in \ch{SrTiO3} with $Z=8$ and despite the smaller relative combinatorial search space, make the prediction of a high-quality structure more difficult. Apart from these, we observe that the ROG curve of our method is always below or intersecting with the IPCSP one, which in turn is always below both classical methods. Finally, the ROG of the structure predicted by our method for \ch{Ca3Al2Si3O12} is slightly lower than the one for \ch{SrTiO3} with $Z=27$, despite the larger relative combinatorial search space size, as shown by the red points on the far-right of the figure. This happens because of the highly repetitive nature of \ch{SrTiO3} with $Z=27$ that introduces additional difficulty in escaping local minima. 

\begin{figure}[ht!]
    \centering
    \includegraphics[width=0.5\linewidth]{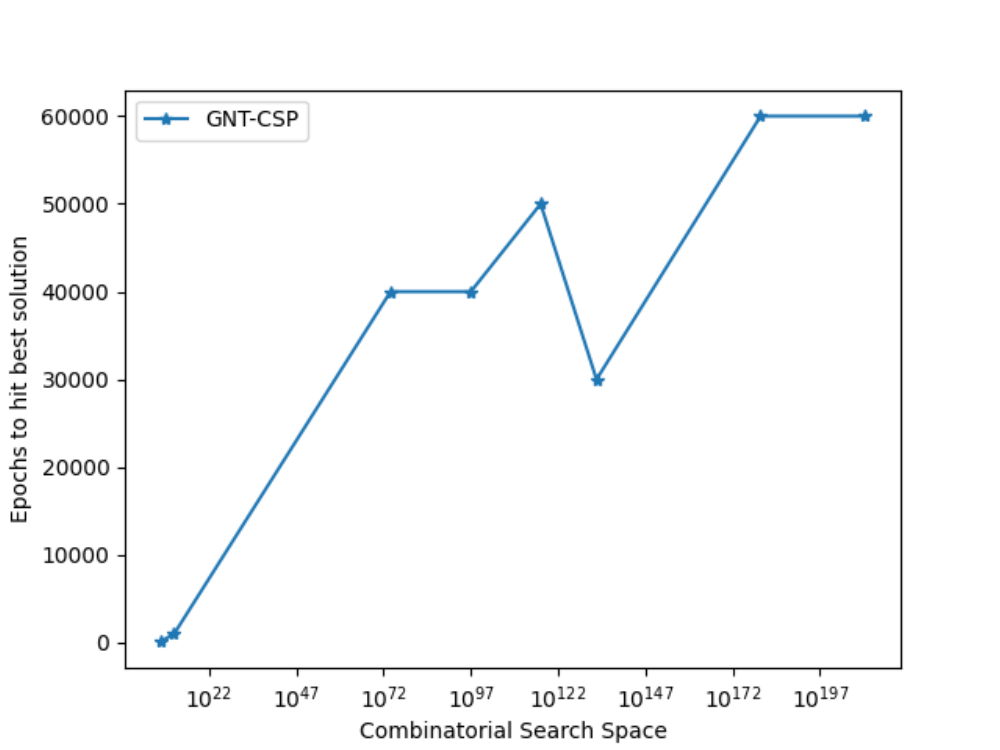}
    \caption{\textbf{Number of epochs our method needed to compute the lowest energy structure of each composition as a function of their combinatorial search space (in logarithmic scale).} Our model tends to need more training epochs to find low-energy feasible structures as the the combinatorial search space of a composition gets larger, while inherent symmetries found in \ch{SrTiO3} with $Z=\{8,27\}$ require a comparable number of epochs to compositions with a slightly larger combinatorial search space.}
    \label{fig:rog_epochs}
\end{figure}

We also provide, in Figure~\ref{fig:rog_epochs}, the number of training epochs that our approach requires to predict the best feasible structure as a function of the size of the combinatorial search space of the compositions we examined. Intuitively, our method needs more epoch as the size of the combinatorial search space of compositions grows. In a similar fashion to ROG, the epochs' number is also impacted by the inherent symmetries of the investigated compositions, with \ch{SrTiO3} ($Z=8$) and \ch{SrTiO3} ($Z=27$) needing almost as many epochs as compositions with slightly larger combinatorial search spaces.

We see a practical application of our method in materials discovery workflows as a high-quality generator of probe structures capable of leveraging GPU infrastructure~\cite{Collins2017, doi:10.1021/acs.accounts.4c00694}. Probe structures do not necessarily provide the absolute global energy minimum for a given composition, but rather an approximation whose primary purpose is to guide experimental synthesis decisions. As a chemist selects a chemical system to study, existing compounds define the convex hull, representing the energy threshold for every composition. Probe structures with energies below the convex hull indicate the potential presence of previously unknown compounds within a given chemical system. In this context, the aim of a structure search method is not to find exact structures at specific compositions, but to identify high-quality probe structures on average across the chemical space. Notably, in this work we do not fully assess the performance of our method in this setting, but instead focus specifically on the combinatorial aspect of CSP, omitting the usual sources of improvement such as structure relaxation, symmetry constraints, and related refinements.

Overall, GNT-CSP consistently finds high-quality atom allocations for a range of compositions that have combinatorial search spaces of different magnitudes and involve different levels of symmetries. The very good and promising performance of our method demonstrates its applicability and sets the ground for future use in larger scale experiments and for additional compositions.

\begin{table}[ht]
\begin{center}
\begin{tabular}{|c c c c c c c|} 
 \hline
Composition & Metrics & Ground truth & Greedy & SA & IPCSP & GNT-CSP  \\
    \hline
  \ch{SrTiO3}
  & ROG & 0 & \textbf{0} & \textbf{0} & \textbf{0} & \textbf{0}\\
 g=4, Z=1 & Energy & -158.76 & \textbf{-158.76} & \textbf{-158.76} & \textbf{-158.76} & \textbf{-158.76}\\
    \hline
  \ch{SrTiO3}
   & ROG & 0 & \textbf{0} & \textbf{0} & \textbf{0} & \textbf{0} \\
 g=8, Z=1 & Energy & -158.76 & \textbf{-158.76} & \textbf{-158.76} & \textbf{-158.76} & \textbf{-158.76}  \\
  \hline
  \ch{SrTiO3},
  & ROG & 0 & 0.12 & 0.11 & 0.07 & \textbf{0} \\
 Z=8 & Energy & -1268.67 & -1121.39 & -1129.15 & -1179.67 & \textbf{-1268.67}\\
  \hline
 \ch{MgAl2O4}
  & ROG & 0 & 0.09 & 0.08 & 0.06 & \textbf{0.05}\\
  & Energy& -1620.89 & -1483.11 & -1486.06 & -1525.63 & \textbf{-1546.48} \\
    \hline
 \ch{Y2O3} 
   & ROG & 0 & 0.18 & $0.12$ & \textbf{0.04} & 0.04\\
  & Energy & -2191.57 & -1795.61 & -1932.94 & \textbf{-2111.54} & -2096.51 \\
  \hline
\ch{Y2Ti2O7} 
   &  ROG & 0 & 0.14 & 0.13 & \textbf{0.05} & 0.05\\
   & Energy & -3093.53 & -2660.99 & -2706.28 & \textbf{-2949.29} & -2947.37 \\
    \hline
\ch{SrTiO3},
   & ROG & 0 & 0.16 & 0.15 & \textbf{0.13} & 0.13\\
 Z=27 & Energy & -4281.76  & -3609.5 & -3618.77 & \textbf{-3745.89} & -3742.86 \\
  \hline
 \ch{Ca3Al2Si3O12}
  & ROG & 0 & 0.16 & 0.16 & 0.15 & \textbf{0.10} \\
 & Energy & -2199.99 & -1840.75 & -1852.75 & -1880.42 & \textbf{-1986.79}\\
    \hline
 \end{tabular}
 \caption{ROG and Energy values (in eV) of the best structures computed by each method and for each composition.}
 \label{tab:main_results}
\end{center}
\end{table}

\section{Conclusions and Future Work}
In this work, we introduce a neural combinatorial approach to the CSP problem that intrinsically accounts for its constraints, without requiring penalty terms or precomputed solutions. This is achieved by projecting node embeddings computed by a graph neural network into the feasible solution space and directly sampling low-energy structures. We additionally introduce an expander-inspired 3D graph construction that consists of both local and long-range connections, which compensates for the locality shortcomings in our case of, widely used in the literature, radius cutoff graphs. The competitive, and sometimes superior, performance of our approach with respect to classical heuristics and a reference solver, as evidenced by our experiments on multiple compositions with varying levels of symmetry and combinatorial search space sizes, paves the way for large scale crystal structure prediction tasks, where traditional solvers tend to struggle due to the very high dimensionality. Its ability to take advantage of GPU infrastructure enables effective materials discovery campaigns centered around extensive probe structure calculations. Subsequent improvements in the design of communication graphs and the integration of a wider range of potential energy functions can further improve the efficiency and applicability of our approach to structure search problems.


\section*{Acknowledgements}
We thank Helmut Katzgraber for his genuine support and vision for interdisciplinary research. This work was supported by a studentship from the School of Electrical Engineering, Electronics and Computer Science, at the University of Liverpool,  UK.  V. V. Gusev is supported by the Leverhulme Research Fellowship and UK Engineering and Physical Sciences Research Council (EPSRC) through AI for Chemistry: AIchemy Hub (EP/Y028775/1 and EP/Y028759/1).

\bibliographystyle{unsrtnat}
\bibliography{references}  
\clearpage

\appendix
\section{Matrix scaling problem and Sinkhorn-Knopp algorithm} ~\label{matrix scaling-sinkhorn}
The matrix scaling problem looks for positive multipliers for the rows and columns of a non-negative matrix that can scale it to have prescribed row and column sums. Formally, given $\mathbf{A}\in \mathbb{R}^{n \times m} \ge 0$, positive target row sums $\mathbf{r} = [r_1, \dots , r_n ]>0$ and positive target column sums $\mathbf{c}=[ c_1, \dots , c_m]>0$, the goal of the matrix scaling problem is to obtain positive diagonal matrices $\mathbf{P}=\operatorname{diag}(\rho_1,\dots,\rho_n)$, $\mathbf{\Gamma}=\operatorname{diag}(\gamma_1,\dots,\gamma_m)$ such that the scaled matrix $\mathbf{X}=\mathbf{P}\mathbf{A}\mathbf{\Gamma}$ has row sums $\mathbf{r}$ and column sums $\mathbf{c}$ : 
\begin{equation*}
 \mathbf{P}\mathbf{A}\mathbf{\Gamma}\mathbf{1}_n=\mathbf{r}, \quad  (\mathbf{P}\mathbf{A}\mathbf{\Gamma})^{\top}\mathbf{1}_n=\mathbf{c},
\end{equation*} 
where $\mathbf{1}_k=(1, \dots,1)^{\top}$ is a k-dimensional vector full of ones. A necessary condition to solve this problem is that the sum of all target sums is equal to the target sum of all row sums, $\sum_{i=1}^{n} r_i = \sum_{j=1}^{m} c_j$.

Mathematically, after initializing $\mathbf{A}^{(0)}=\mathbf{A}$, the Sinkhorn-Knopp algorithm~\cite{83900530-3d01-3dd8-a11c-d939e82fce35} is defined as follows and implemented for a fixed number of $K$ steps:
\begin{equation}
\begin{aligned}
\text{For $k=0,1, \dots, K-1$ }: \\
\text{If } k \text{ is even, row scaling :} \quad 
& \mathbf{A}_{ij}^{(k+1)} = \frac{\mathbf{A}_{ij}^{(k)} \mathbf{r}_i}{\sum_{j} \mathbf{A}_{ij}^{(k)}}, \forall\ i \in \{1,\dots,n\}, \\[6pt]
\text{If } k \text{ is odd, column scaling:} \quad
& \mathbf{A}_{ij}^{(k+1)} = \frac{\mathbf{A}_{ij}^{(k)} \mathbf{c}_j}{\sum_{i} \mathbf{A}_{ij}^{(k)}},\forall\ j \in \{1,\dots,m\}
\end{aligned}
\label{eq:generalized_sinkhorn}
\end{equation}

\section{Entropic regularization} ~\label{entropic}
Theorem 1 shows that we can obtain a basic feasible solution $\operatorname{BFS(\mathbf{H})}$ that lies on a vertex of the transportation polytope $\mathcal{U}(\mathbf{r},\mathbf{c})$, as the limit of $S(\mathbf{H}/\tau)$. This means that we can approximate $\operatorname{BFS(\mathbf{H})} \approx S(\mathbf{H}/\tau)$, for a small value of $\tau$.

\textbf{Theorem 1}. \textit{For a solution $\mathbf{B}$ that belongs to the transportation polytope $\mathcal{U}(\mathbf{r},\mathbf{c})$, define its entropy as $h(\mathbf{B})=-\sum_{i,j}B_{i,j}\mathrm{log}B_{ij}$. Then, }
\begin{equation}
    S(\mathbf{H}/ \tau) = \argmax_{\mathbf{B}\in \operatorname{}\mathcal{U}(\mathbf{r},\mathbf{c})} \langle \mathbf{B} , \mathbf{H} \rangle_F\ + \tau h(\mathbf{B}).
\label{eq:theorem}
\end{equation}
\textit{Assume further that the entries of $\mathbf{H}$ are drawn from a distribution absolutely continuous with respect to the Lebesgue measure on $\mathbb{R}$. Then, it almost surely holds that:}
\begin{align}
    BFS(\mathbf{H}) = \lim_{\tau \to 0^+} S(\mathbf{H} / \tau)
\end{align}

In practice, an incomplete version of the Sinkhorn operator that was described earlier is applied. The number of Sinkhorn iterations directly depends on $\tau$, with smaller $\tau$ values requiring more Sinkhorn iterations to achieve convergence to a vertex of $\mathcal{U}(\mathbf{r},\mathbf{c})$. 

\section{Bottlenecks in graphs}

\begin{figure}[H]
    \centering
\begin{tikzpicture}[
    every node/.style={circle, draw, thick, inner sep=2pt},
    blueNode/.style={fill=blue!20},
    purpleNode/.style={fill=purple!20}
]

\node[blueNode] (a1) at (0,2) {};
\node[blueNode] (a2) at (1.2,2) {};
\node[blueNode] (a3) at (2.4,2) {};
\node[blueNode] (a4) at (0,0) {};
\node[blueNode] (a5) at (1.2,0) {};
\node[blueNode] (a6) at (2.4,0) {};

\foreach \i [count=\ii from 1] in {a1,a2,a3,a4,a5,a6}{
  \foreach \j [count=\jj from 1] in {a1,a2,a3,a4,a5,a6}{
    \ifnum\ii<\jj
      \draw (\i) -- (\j);
    \fi
  }
}

\node[purpleNode] (b1) at (5,2) {};
\node[purpleNode] (b2) at (6.2,2) {};
\node[purpleNode] (b3) at (7.4,2) {};
\node[purpleNode] (b4) at (5,0) {};
\node[purpleNode] (b5) at (6.2,0) {};
\node[purpleNode] (b6) at (7.4,0) {};

\foreach \i [count=\ii from 1] in {b1,b2,b3,b4,b5,b6}{
  \foreach \j [count=\jj from 1] in {b1,b2,b3,b4,b5,b6}{
    \ifnum\ii<\jj
      \draw (\i) -- (\j);
    \fi
  }
}

\draw[green, thick] (a6) -- (b4);   

\end{tikzpicture}
\caption{\textbf{Barbell graph.} One typical example of a bottleneck that can lead to oversquashing can be found in a barbell graph, where the green edge is responsible to propagate information between two large groups of nodes.}
\label{fig:barbell}
\end{figure}
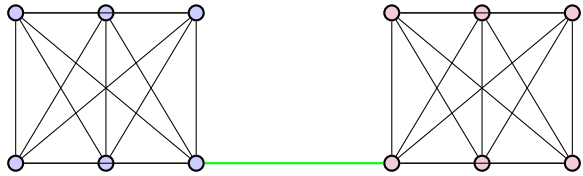

\section{Architecture of our model}

\tikzset{
  block/.style={
    draw, rounded corners,
    minimum width=2.2cm,
    minimum height=0.8cm,
    align=center
  },
  plus/.style={
    circle, draw,
    inner sep=1pt
  },
  arrow/.style={->, thick}
}

\begin{figure}[H]
\centering
\resizebox{\linewidth}{!}{%
\begin{tikzpicture}[
    font=\small,
    block/.style={
        draw,
        rounded corners=2pt,
        minimum width=2.8cm,
        minimum height=0.9cm,
        align=center,
        line width=0.8pt,
        fill=blue!15
    },
    inter/.style={
        draw,
        rounded corners=2pt,
        minimum width=2.8cm,
        minimum height=0.65cm,
        align=center,
        line width=0.8pt,
        fill=blue!30
    },
    arrow/.style={->, line width=0.8pt},
    dashbox/.style={
        draw,
        dashed,
        line width=0.9pt,
        rounded corners=3pt,
        inner sep=6pt
    }
]

\node[block] (embed) {Embedding};
\node[block, right=6mm of embed] (linear1) {Linear};

\node[inter, below=12mm of $(embed)!0.5!(linear1)$] (int1) {GINE Layer};
\node[inter, below=4mm of int1] (int2) {GINE Layer};
\node[inter, below=4mm of int2] (int3) {GINE Layer};

\node[below=5mm of int3] (output) {Output};

\draw[->, line width=0.8pt]
  ($(embed.north)+(0,1)$) --
  node[midway, above, yshift=5.0mm, align=center]
  {$n$}
  (embed.north);
  
\draw[->, line width=0.8pt]
  ($(linear1.north)+(0,1)$) --
  node[midway, above, yshift=5.0mm, align=center]
  {$\mathbf{Q}_i^c$}
  (linear1.north);

\draw[arrow] (embed.south) -- 
  node[midway, above, xshift=3.0mm, yshift=-2.0mm, align=center]
  {$d_0$}
(int1.north west);

\draw[arrow] (linear1.south) -- 
  node[midway, above, xshift=3.0mm, yshift=-2.0mm, align=center]
  {$b_0$}
(int1.north east);

\draw[arrow] (int1) -- (int2);
\draw[arrow] (int2) -- (int3);
\draw[arrow] (int3) -- (output);

\node[right=4.2cm of int2] (hv_top) {$\mathbf{h}_v$};
\node[right=3.2cm of hv_top] (hj) {$\mathbf{h}_j$};
\node[right=3.2cm of hj] (ejv) {$\mathbf{e}_{jv}$};

\node[block, below=6mm of ejv] (linear) {Linear};

\node[plus, below=12mm of hj] (plus1) {$+$};

\draw[arrow] (hj) -- (plus1);
\draw[arrow] (ejv) -- (linear);
\draw[arrow] (linear.west) -- ++(-1.4,0) |- (plus1);

\node[block, below=8mm of plus1] (gelu1) {GeLU};
\node[below=5mm of gelu1] (zj) {$\sum_j$};
\node[below=10mm of hv_top] (alpha) {1+$\alpha$};

\draw[arrow] (hv_top) -- (alpha);
\draw[arrow] (plus1) -- (gelu1);
\draw[arrow] (gelu1) -- (zj);

\node[plus, below=12mm of zj] (plus2) {$+$};

\draw[arrow] (zj) -- (plus2);
\draw[arrow] (alpha) |- node[midway, left] {$(1+\alpha)\mathbf{h}_v$} (plus2);

\node[block, below=8mm of plus2] (mlp) {MLP};
\node[block, below=6mm of mlp] (gelu2) {GeLU};
\node[below=6mm of gelu2] (hv_bot) {$\mathbf{h}_v$};

\draw[arrow] (plus2) -- (mlp);
\draw[arrow] (mlp) -- (gelu2);
\draw[arrow] (gelu2) -- (hv_bot);

\node[dashbox,
  fit=(hv_top)
    (hj)
    (ejv)
    (linear)
    (plus1)
    (gelu1)
    (zj)
    (plus2)
    (mlp)
    (gelu2)
    (hv_bot)
  ,
  label={[rotate=90]right:GINE layer Block}
] {};

\draw[arrow, dashed]
  (int2.east)
  .. controls +(2.2,0.8) and +(-2.2,0) ..
  (hv_top.west);
  
\end{tikzpicture}}
\caption{\textbf{The developed model architecture for our GNN.} An embedding block is used to create a $d_0$-dimensional node embedding for each of the $n$ positions. We create $b_0$-dimensional edge embeddings by taking the respective columns of the interactions matrix $\mathbf{Q}$ and passing them from a linear layer. Node and edge embeddings are fed to GINE layers which generate node embeddings according to message-passing as shown in the GINE layer block. The output of the last GINE layer provides the final node $k+1$-dimensional node embeddings. }
\label{fig:gnn_model}
\end{figure}
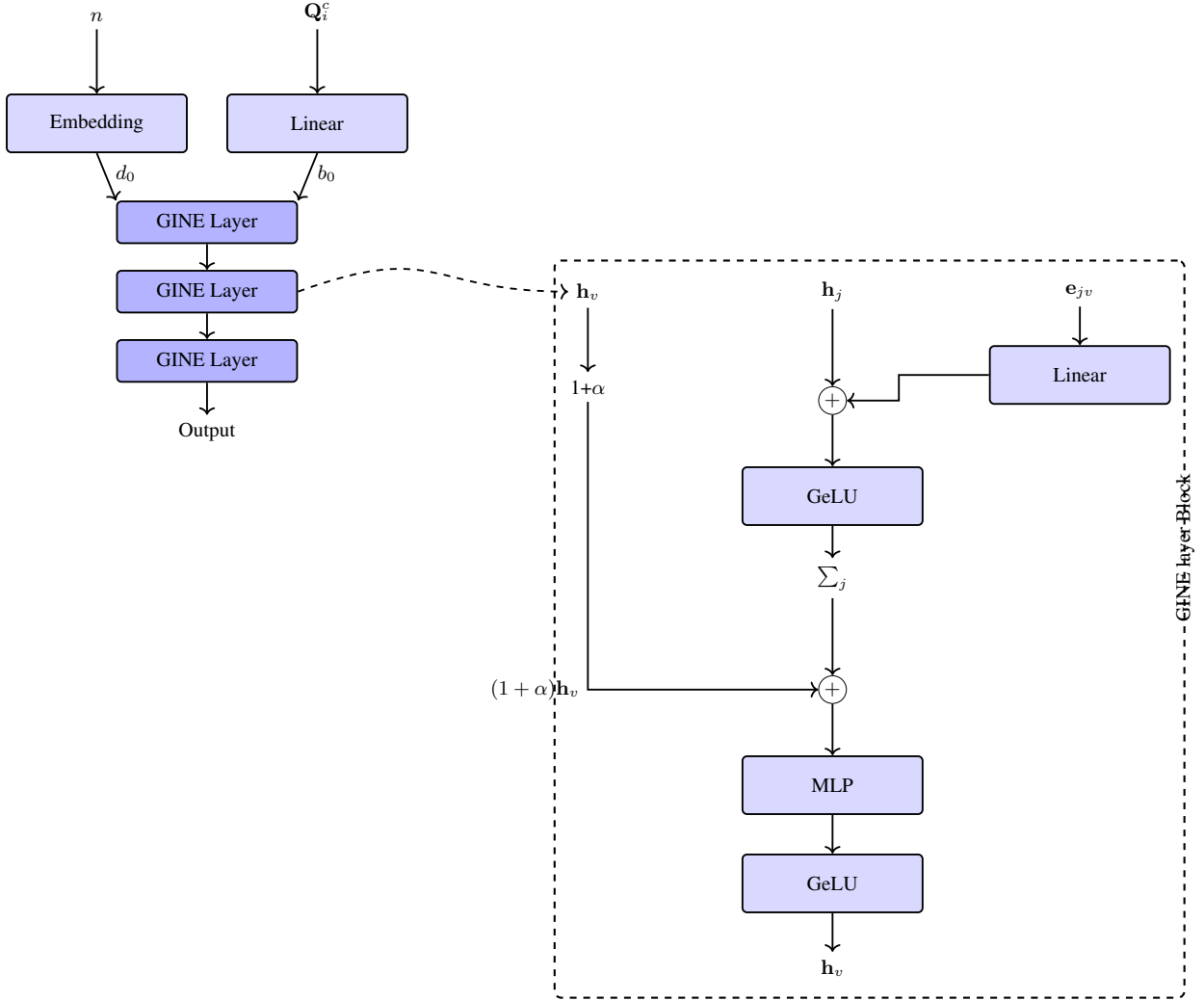

\section{Radius cutoff and margulis graphs} \label{radius_margulis}
To highlight the gain of introducing the 3D Gabber-Gallil graph, we additionally conducted experiments on radius cutoff graphs as well as  Margulis 3D graphs which we created following the same logic as the Gabber-Galil 3D construction. Importantly, the topologies of both Margulis 3D and Gabber-Galil 3D change only as a result of altering the number of nodes, that is controlled by the discretization parameter $g$. Therefore, the same graph is created for all compositions that share the same $g$. However, this is not the case for the radius cutoff graphs, whose construction depends on the radius cutoff threshold and the cell parameter of each composition. For the following experiments, we empirically selected  a radius cutoff threshold of 4 \AA{} and 16 nearest neighbors.

\begin{figure}[h]
\centering

\begin{subfigure}[b]{0.30\textwidth}
  \includegraphics[width=\linewidth]{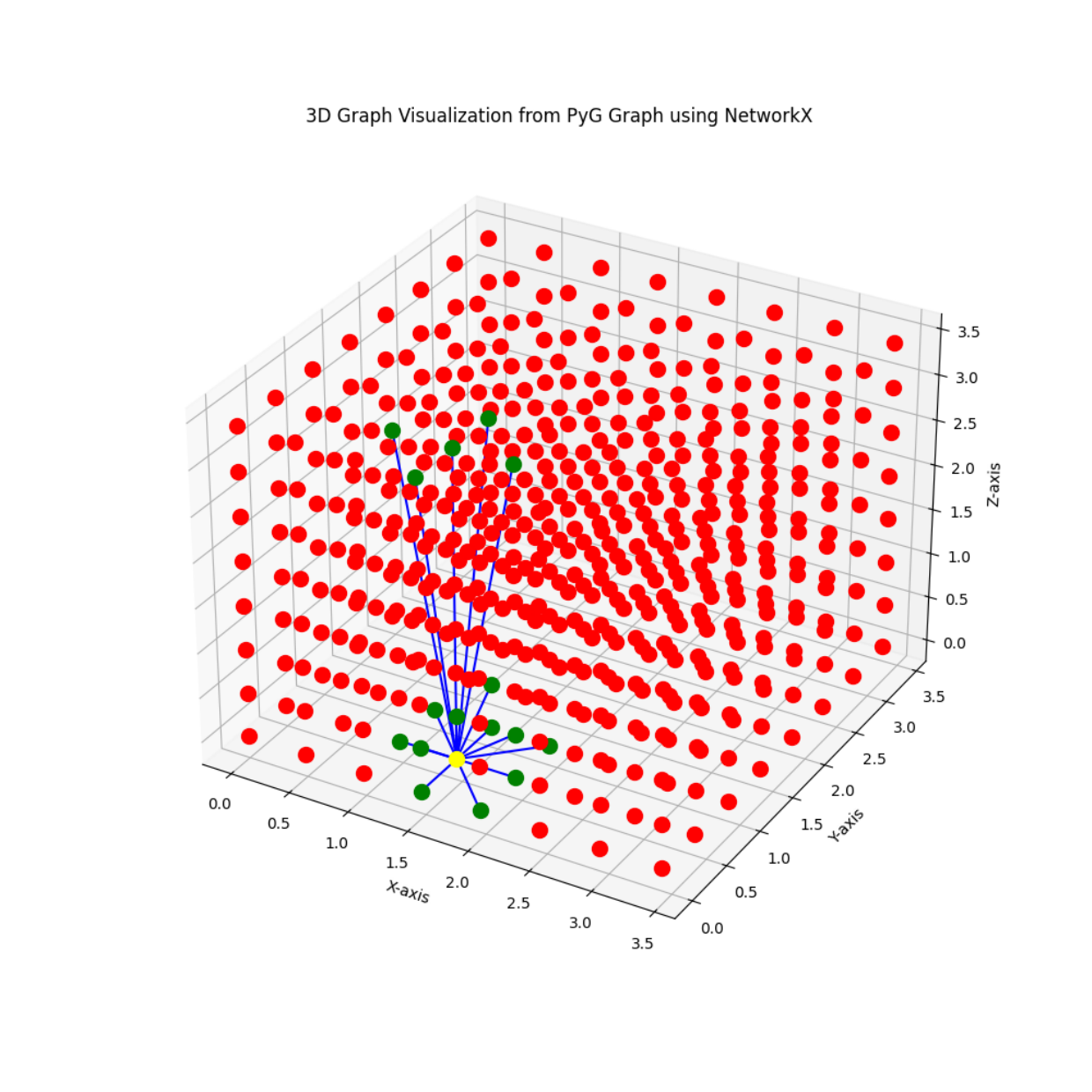}
  \caption{Cutoff}
  \label{fig:graphs_1a}
\end{subfigure}
\begin{subfigure}[b]{0.30\textwidth}
  \includegraphics[width=\linewidth]{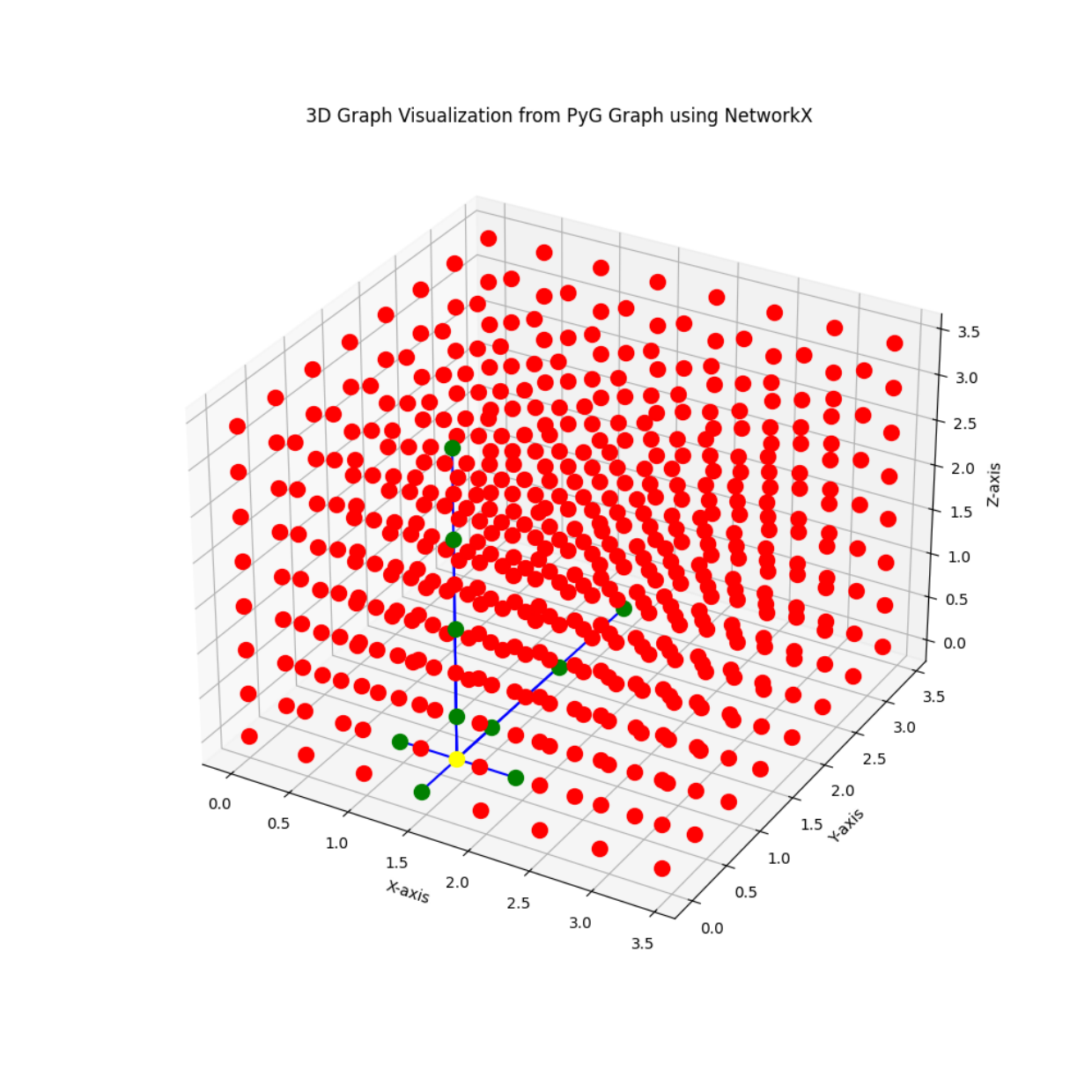}
  \caption{Margulis 3D}
  \label{fig:graphs_1b}
\end{subfigure}
\begin{subfigure}[b]{0.30\textwidth}
  \includegraphics[width=\linewidth]{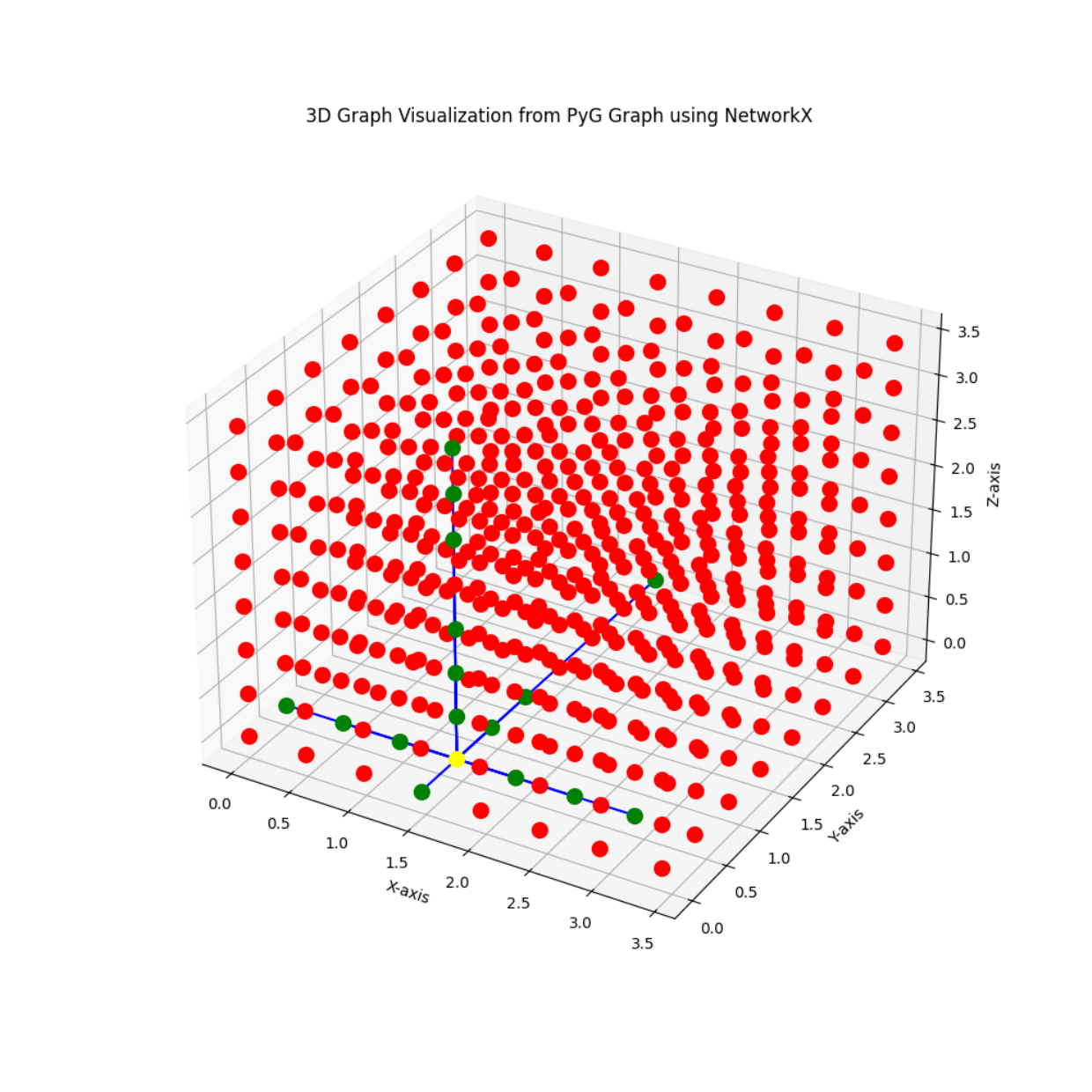}
  \caption{Gabber-Galil 3D}
  \label{fig:graphs_1c}
\end{subfigure}

\begin{subfigure}[b]{0.30\textwidth}
  \includegraphics[width=\linewidth]{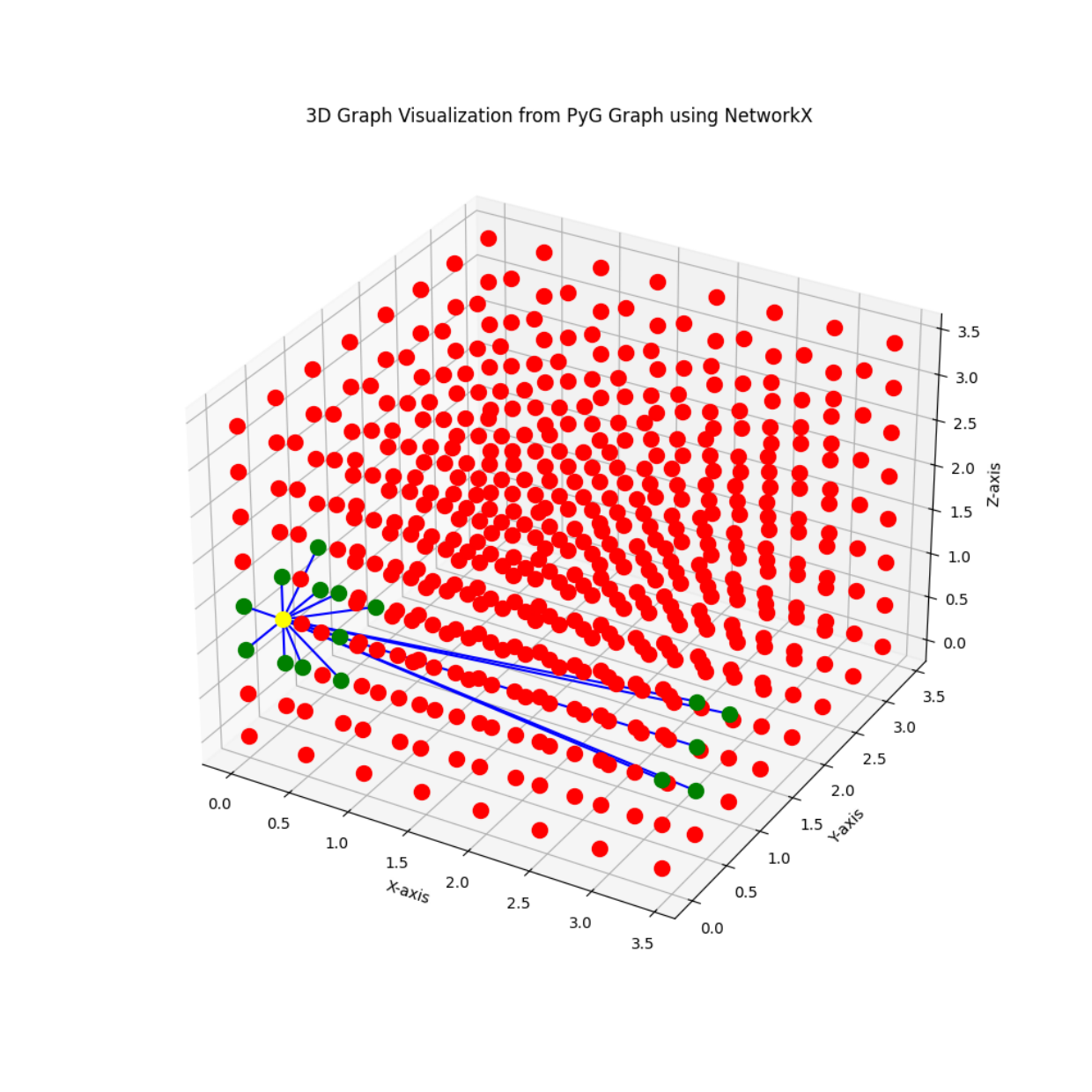}
  \caption{Cutoff}
  \label{fig:graphs_3a}
\end{subfigure}
\begin{subfigure}[b]{0.30\textwidth}
  \includegraphics[width=\linewidth]{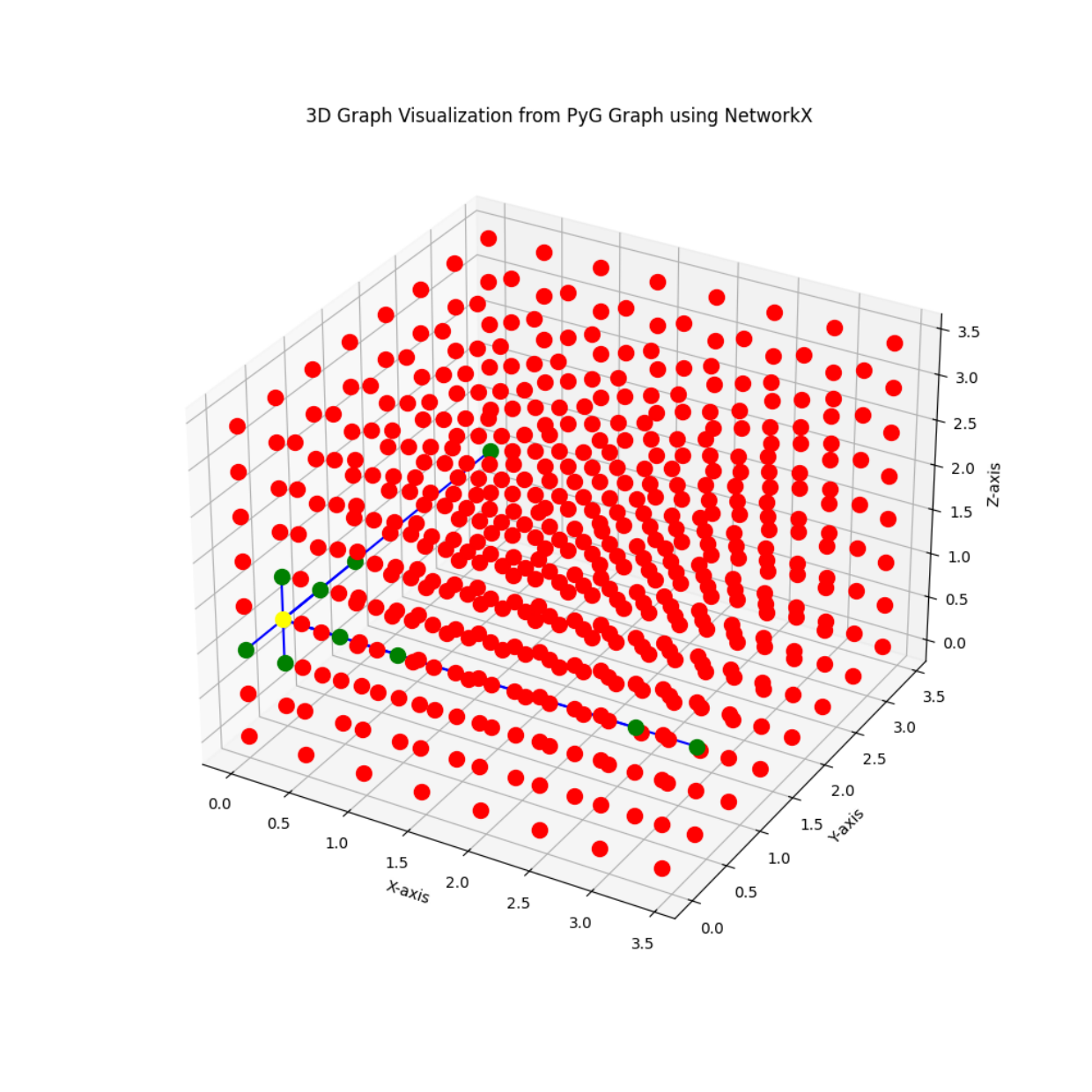}
  \caption{Margulis 3D}
  \label{fig:graphs_3b}
\end{subfigure}
\begin{subfigure}[b]{0.30\textwidth}
  \includegraphics[width=\linewidth]{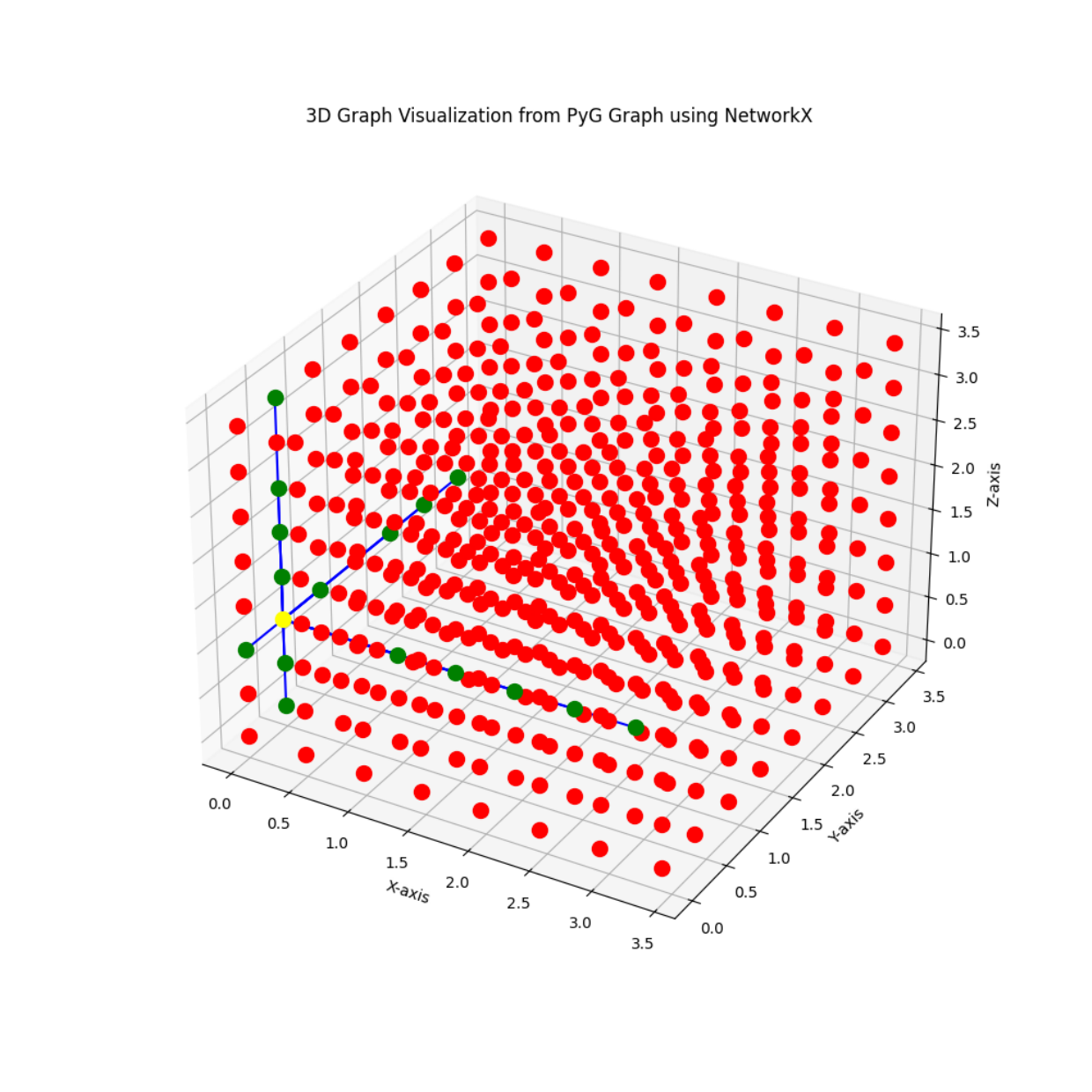}
  \caption{Gabber-Galil 3D}
  \label{fig:graphs_3c}
\end{subfigure}

\begin{subfigure}[b]{0.30\textwidth}
  \includegraphics[width=\linewidth]{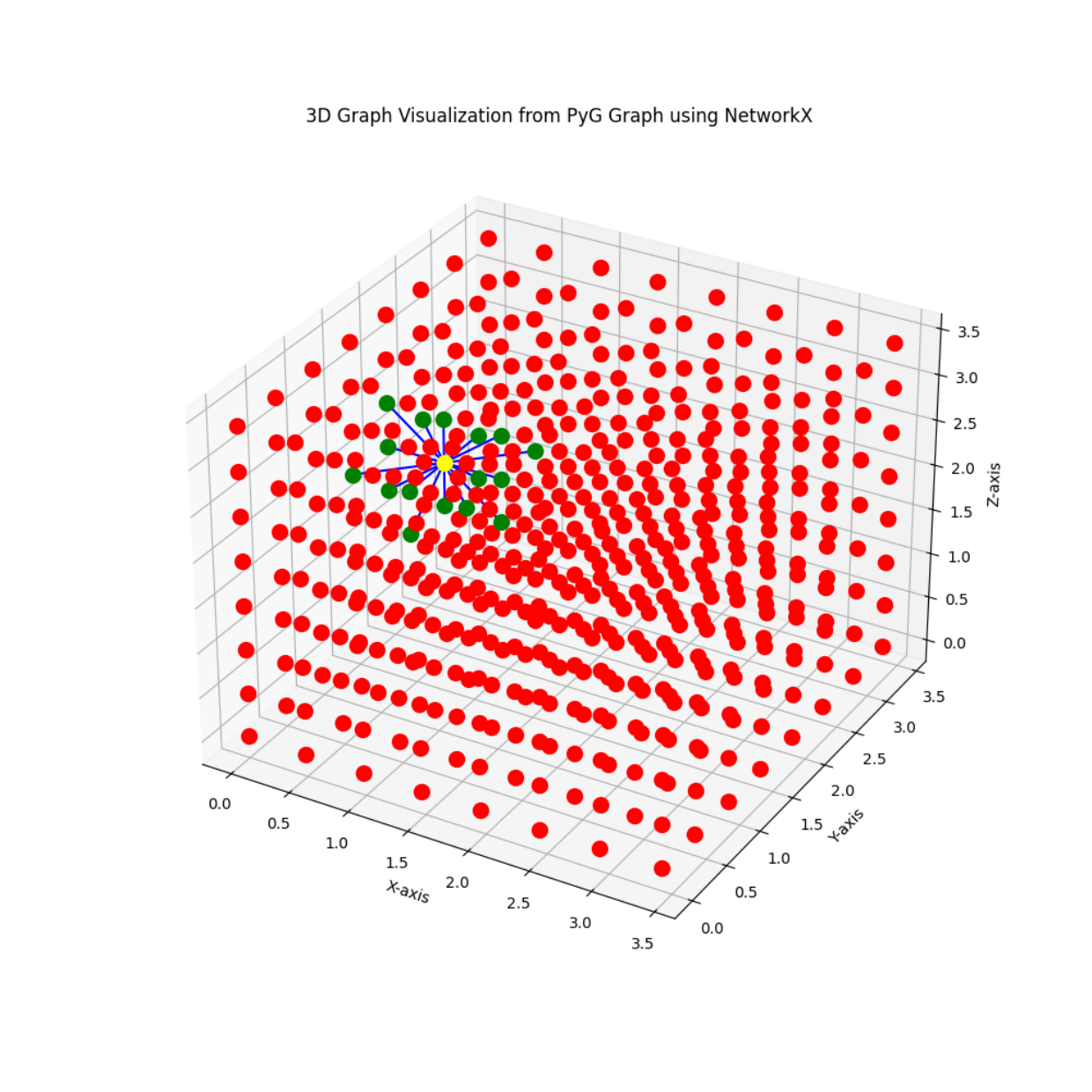}
  \caption{Cutoff}
  \label{fig:graphs_4a}
\end{subfigure}
\begin{subfigure}[b]{0.30\textwidth}
  \includegraphics[width=\linewidth]{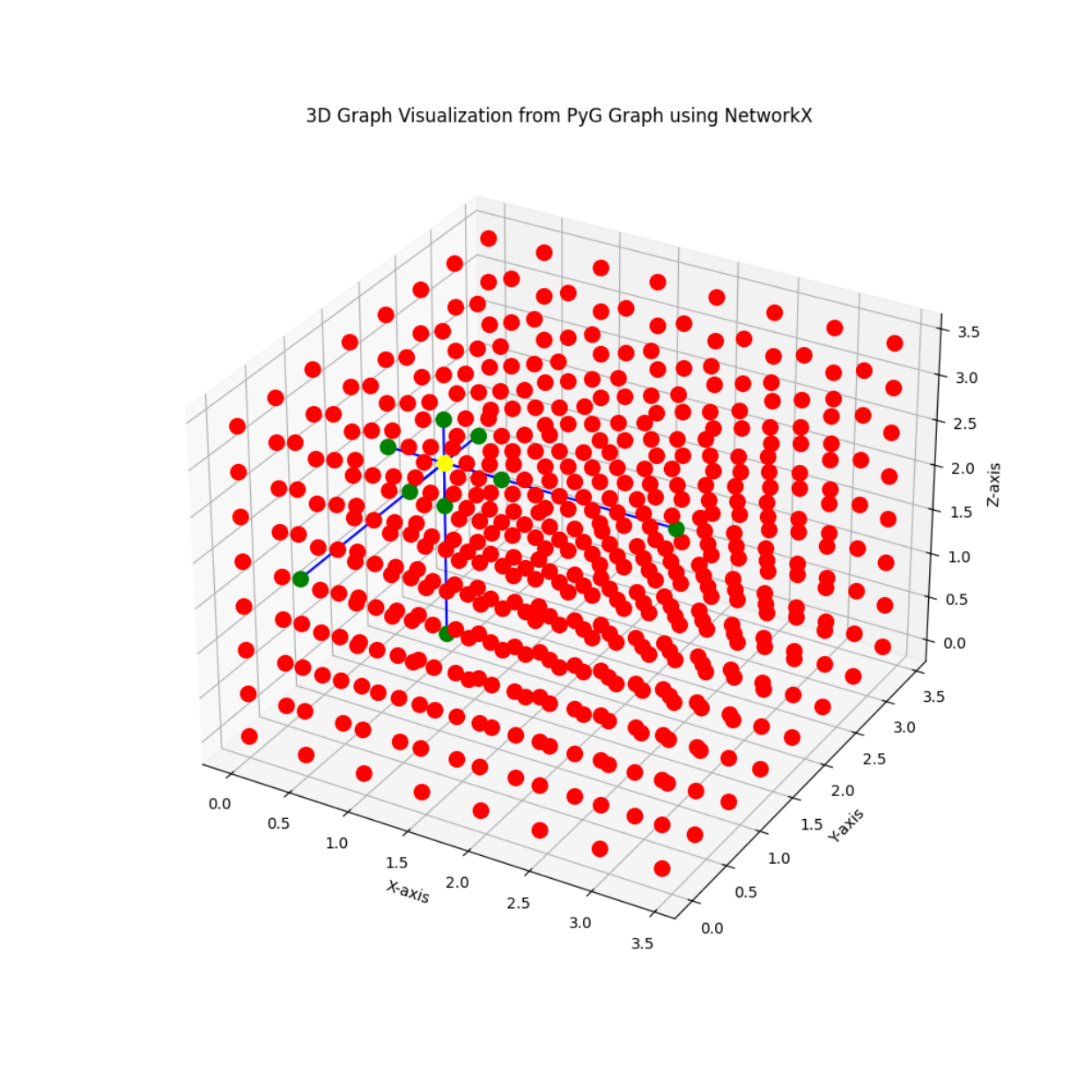}
  \caption{Margulis 3D}
  \label{fig:graphs_4b}
\end{subfigure}
\begin{subfigure}[b]{0.30\textwidth}
  \includegraphics[width=\linewidth]{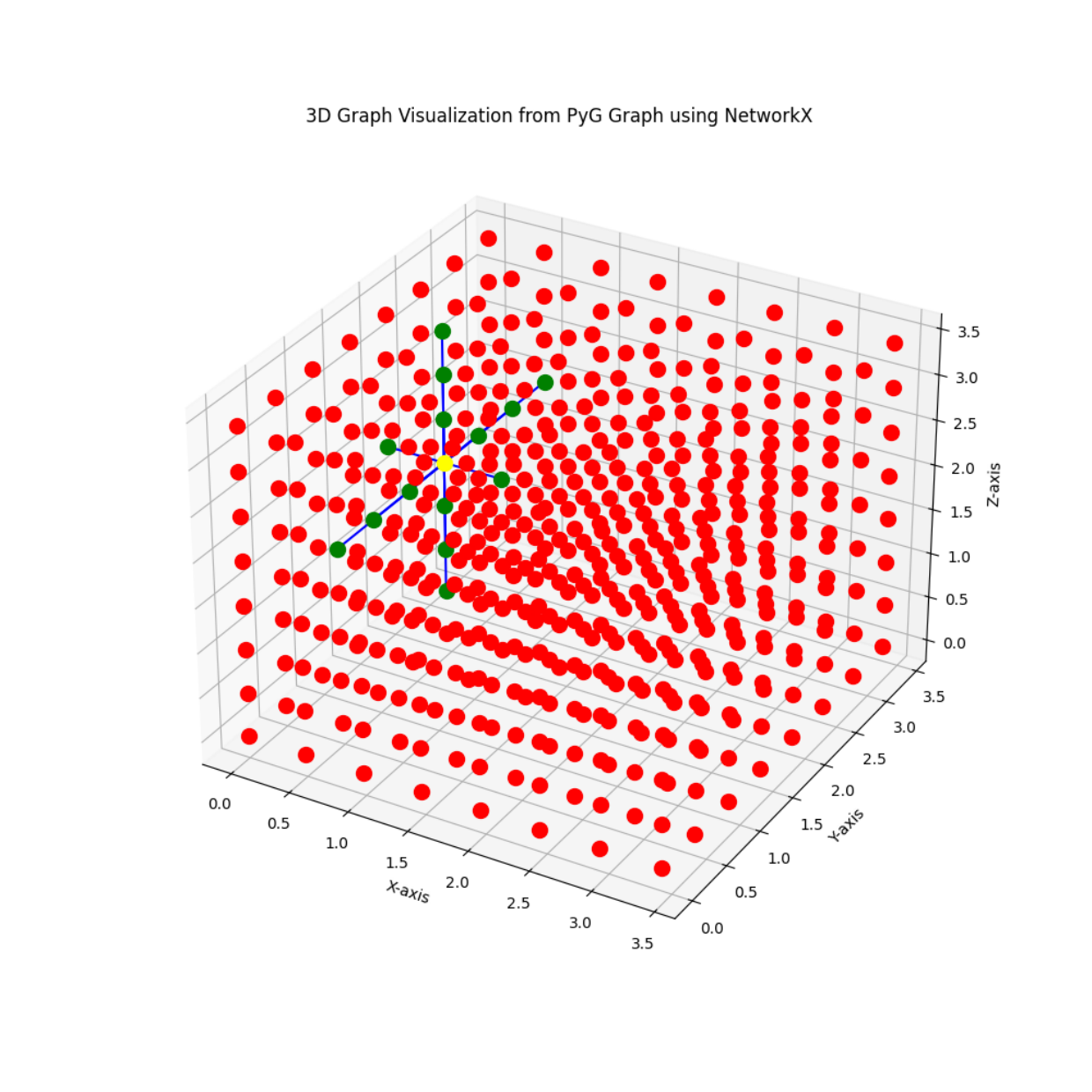}
  \caption{Gabber-Galil 3D}
  \label{fig:graphs_4c}
\end{subfigure}
\caption{\textbf{Instances of one-hop node connectivities for radius cutoff, Margulis 3D and Gabber-Galil 3D graphs.} Each row of subfigures illustrates the connections (blue edges) of the same central node (in yellow color) with its one-hop neighbors (in green color) for all three graphs. Radius cutoff graphs capture mainly connections between neighboring nodes, while periodicity introduces few connections between more distant nodes. Margulis 3D captures both local and long-range connections but its expansion seems limited compared to Gabber-Galil 3D which includes long-range connections across all directions. }
\label{fig:graph_plots}
\end{figure}

We provide a few instances of node connectivities for all three types of graphs in Figure~\ref{fig:graph_plots}. As expected, radius cutoff graphs, instances of which are shown in figures~\ref{fig:graphs_1a},~\ref{fig:graphs_3a},~\ref{fig:graphs_4a}, favor connections between neighboring nodes, while they also incorporate a small number of long-range interactions that account for periodicity. Margulis 3D graphs, whose connectivities are shown in figures~\ref{fig:graphs_1b},~\ref{fig:graphs_3b},~\ref{fig:graphs_4b}, are designed to include interactions between both neighboring and distant nodes. Nevertheless, Gabber-Galil 3D graphs, illustrated in figures~\ref{fig:graphs_1c},~\ref{fig:graphs_3c},~\ref{fig:graphs_4c}, expand better than the Margulis 3D graphs as they capture long-range interactions across all directions.

Table~\ref{tab:results_per_graph} provides ROG and energy values of the structures computed by our method when using different graph types and with respect to each composition. Our method, when applied with radius-cutoff graphs, finds solutions that are comparable to SA (slightly better or worse) but definitely worse than IPCSP in terms of energy, except for \ch{SrTiO3} with $Z=27$. When using the Margulis 3D graphs, our method computes comparable solutions in terms of energy and ROG with the solutions associated with the radius cutoff graphs. More specifically, our method with Margulis 3D graphs provides better solutions than with radius cutoff graphs in the cases of \ch{SrTiO3} with $Z=8$ and \ch{Y2O3} and worse solutions in the rest of the compositions.
However, when using the Gabber-Galil 3D graph, our method samples higher quality solutions than both radius cutoff graphs and 3D margulis graphs since the ROG drops and the energy gets closer to the ground truth one. We argue that this happens because of the expander nature of the Gabber-Galil 3D graphs that include both local and long-range connections, compared to the radius cutoff graphs that favor local connections by design and Margulis 3D graphs which have less powerful expansion properties. As a result, by performing the message-passing on 3D Gabber-Galil graphs the GNN can learn more powerful node representations that subsequently improve the quality of the obtained solutions. 

\begin{figure}
    \centering
    \includegraphics[width=0.5\linewidth]{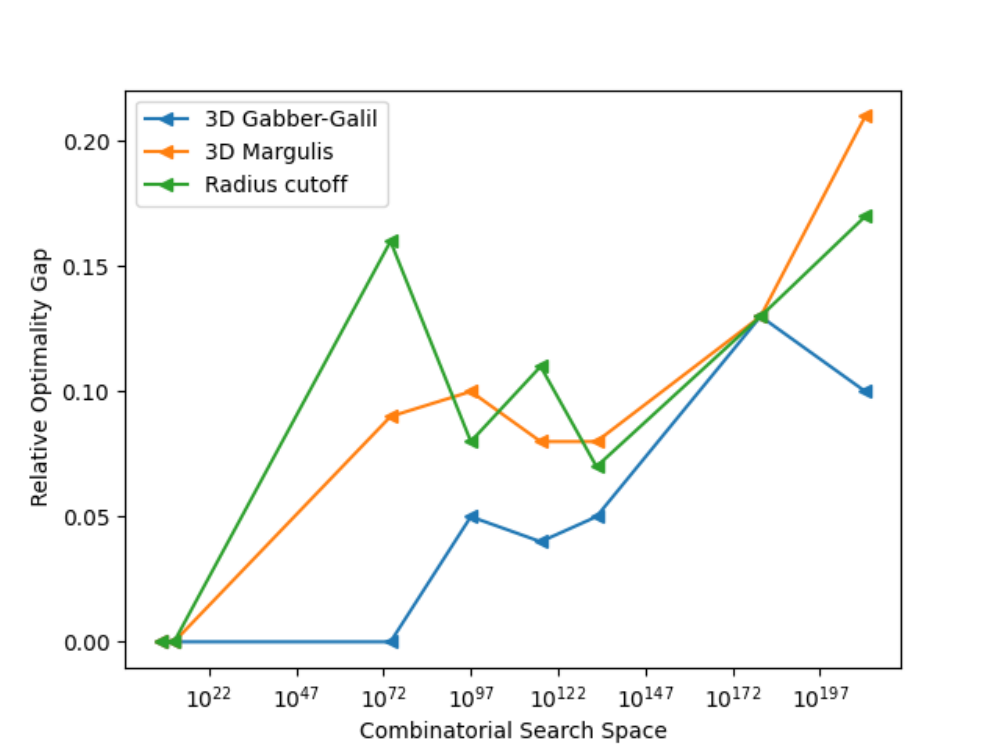}
    \caption{\textbf{ROG curves for the structures GNT-CSP computed with respect to type of graph we performed the message-passing on and as a function of the combinatorial search space, in logarithmic scale, of the compositions we experimented with.} The blue curve that corresponds to the case of the Gabber-Galil 3D graph is below both the orange one (3D margulis) and the green one (radius cutoff), indicating the consistently higher-quality solutions that are associated with choosing it.}
    \label{fig:rog_graphs}
\end{figure}

The observations we made above are shown in Figure~\ref{fig:rog_graphs} that illustrates the ROG of the structures our method computed using the different types of graphs as a function of the size of the combinatorial search space of the compositions that were investigated.
\begin{table}[h!]
\begin{center}
\begin{tabular}{|c c c c c c|} 
 \hline
Composition & Metrics & Ground truth & radius-cutoff & Margulis 3D & Gabber-Galil 3D  \\
    \hline
  \ch{SrTiO3} & ROG & 0 & \textbf{0} & \textbf{0} & \textbf{0}\\
  g=4, Z=1& Energy & -158.76 eV & \textbf{-158.76} eV & \textbf{-158.76} eV & \textbf{-158.76} eV\\
    \hline
  \ch{SrTiO3}
  & ROG & 0 & \textbf{0} & \textbf{0} & \textbf{0} \\
  g=8, Z=1 & Energy & -158.76 eV & \textbf{-158.76} eV & \textbf{-158.76} eV & \textbf{-158.76} eV  \\
  \hline
  \ch{SrTiO3}
   & ROG & 0 & 0.16 & 0.09 & \textbf{0} \\
  Z=8 & Energy & -1268.67 eV & -1065.67 eV & -1157.46 eV & \textbf{-1268.67} eV\\
  \hline
 \ch{MgAl2O4}
  & ROG & 0 & 0.08 & 0.10 & \textbf{0.05}\\
  & Energy & -1620.89 eV & -1494.5 eV & -1461.64 eV & \textbf{-1546.48} eV \\
    \hline
 \ch{Y2O3} 
   & ROG & 0 & 0.11 & 0.08 & \textbf{0.04}\\
  & Energy & -2191.57 eV & -1947.82 eV & -2006.32 eV & \textbf{-2096.51} eV \\
  \hline
\ch{Y2Ti2O7} 
   &  ROG & 0 & 0.07 & 0.08 & \textbf{0.05}\\
   & Energy & -3093.53 eV & -2865.87 eV & -2841.92 eV & \textbf{-2947.37} eV \\
    \hline
\ch{SrTiO3} & ROG & 0 & \textbf{0.13} & 0.13 & 0.13\\
 Z=27 & Energy & -4281.76 eV & \textbf{-3746.67} eV & -3702.32 eV &-3742.86 eV\\
  \hline
 \ch{Ca3Al2Si3O12}
  & ROG & 0 & 0.17 & 0.21 & \textbf{0.10} \\
  & Energy & -2199.99 eV & -1806.55 eV & -1743.99 eV & \textbf{-1986.79} eV \\
    \hline
 \end{tabular}
 \caption{ROG and energy values of the best structures our method computed with respect to the graph that was used for message-passing and for every investigated composition.}
 \label{tab:results_per_graph}
\end{center}
\end{table}

\section{Greedy and SA} \label{app:greedy_SA}

SA comes with a few hyperparameters that need to be appropriately selected. These correspond to: the initial temperature $T_0$, the sweep factor $S$ that controls after every how many iterations the temperature is annealed, the cooling factor $a<1$ that controls by how much the temperature is annealed, and the minimum temperature $T_{\min}$.

\begin{algorithm}[H]
\caption{Greedy for CSP solver}
\label{alg:greedy-csp}
\begin{algorithmic}[1]
\REQUIRE Initial configuration $\mathbf{x}_0$, Pairwise interactions matrix $\mathbf{Q}$, number of iterations \texttt{MaxIter}, Set of occupied positions $O$, Set of void positions $V$
\ENSURE Energy of best configuration $\mathbf{x}$, $E(\mathbf{x}) \gets \operatorname{vec}(\mathbf{x})^\top\mathbf{Q}\operatorname{vec}({\mathbf{x}})$

\STATE $\mathbf{x} \gets \mathbf{x}_0$  \COMMENT{Start from a random feasible configuration}
\FOR{$k = 1$ to \texttt{MaxIter}}
    \STATE $p_1\sim O$ \COMMENT{Randomly sample an occupied position $p_1$}
    \STATE $p_2\sim V$ \COMMENT{Randomly sample an empty position $p_2$}
    \STATE $\mathbf{x}' \gets$ Move atom from $p_1$ to $p_2$  \COMMENT{Candidate configuration after constraint-preserving move}
    \STATE $E(\mathbf{x}) \gets \operatorname{vec}(\mathbf{x})^\top\mathbf{Q}\operatorname{vec}({\mathbf{x}})$, $E(\mathbf{x'}) \gets \operatorname{vec}(\mathbf{x'})^\top\mathbf{Q}\operatorname{vec}({\mathbf{x'}})$
    \IF{$E(\mathbf{x}')< E(\mathbf{x]})$}
        \STATE $x \gets x'$  \COMMENT{Accept move if energy lower than current best}
        \STATE $O \leftarrow O \setminus \{p_1\} \cup \{p_2\}$ \COMMENT{Update $O$ to include $p_2$ and exclude $p_1$}
        \STATE $V \leftarrow V \setminus \{p_2\} \cup \{p_1\}$ \COMMENT{Update $V$ to include $p_1$ and exclude $p_2$}
    \ENDIF
\ENDFOR
\RETURN $E(x)$ 
\end{algorithmic}
\end{algorithm}

\begin{algorithm}[H]
\caption{Simulated Annealing for CSP}
\label{alg:sa-csp}
\begin{algorithmic}[1]
\REQUIRE Initial configuration $x_0$ satisfying constraints,  Pairwise interactions matrix $\mathbf{Q}$, initial temperature $T_0$, minimum temperature $T_{\min}$, cooling factor $\alpha$, sweep factor $S$, number of iterations \texttt{MaxIter}, Set of occupied positions $O$, Set of void positions $V$
\ENSURE Energy of best configuration $x_{\text{best}}$, $E(x_{\text{best}})$

\STATE Initialize feasible configuration $\mathbf{x} \gets \mathbf{x}_0$
\STATE $E(\mathbf{x}_{\text{best}}) \gets E(\mathbf{x})$
\STATE $\mathbf{x}_{\text{best}} \gets \mathbf{x}$
\STATE $t \gets T_0$ \COMMENT{Temperature is initialized as $T_0$}

\FOR{k = 1 to \texttt{MaxIter}}
    \STATE $p_1\sim O$ \COMMENT{Randomly sample an occupied position $p_1$}
    \STATE $p_2\sim V$ \COMMENT{Randomly sample an empty position $p_2$}
    \STATE $x' \gets$ Move atom from $p_1$ to $p_2$  \COMMENT{Candidate configuration after constraint-preserving move}
    \STATE $r \sim uniform(0,1)$  \COMMENT{Sample a random number in (0,1)}
    \STATE $E(\mathbf{x}) \gets \operatorname{vec}(\mathbf{x})^\top\mathbf{Q}\operatorname{vec}({\mathbf{x}})$, $E(\mathbf{x'}) \gets \operatorname{vec}(\mathbf{x'})^\top\mathbf{Q}\operatorname{vec}({\mathbf{x'}})$
    \IF{$E(\mathbf{\mathbf{x}}')<E(\mathbf{x})$ \OR $r < \exp((E(\mathbf{x}) - E(\mathbf{x}'))/t)$}
        \STATE $\mathbf{x} \gets \mathbf{x}'$ \COMMENT{Accept move}
        \STATE $O \leftarrow O \setminus \{p_1\} \cup \{p_2\}$ \COMMENT{Update $O$ to include $p_2$ and exclude $p_1$}
        \STATE $V \leftarrow V \setminus \{p_2\} \cup \{p_1\}$ \COMMENT{Update $V$ to include $p_1$ and exclude $p_2$}
        \IF{$E(\mathbf{x}) < E(\mathbf{x}_{\text{best}})$}
            \STATE $\mathbf{x}_{\text{best}} \gets x$ \COMMENT{Update best configuration}
        \ENDIF
    \ENDIF
    
    \IF{$k \bmod (S \cdot n) = 0$}
        \STATE $t \gets \max(t \cdot \alpha, T_{\min})$
    \ENDIF
\ENDFOR

\STATE Return $x_{\text{best}}$
\end{algorithmic}
\end{algorithm}

We performed additional experiments with different strategies to sample new configurations for the Greedy and SA baselines. More specifically, instead of always moving an atom from an occupied position to an empty one, we also tried with some probability to swap atoms of different elements that occupy different positions. Importantly, this strategy also ensures that the new configurations do not violate either of the two constraints. We provide results for three different probability scenarios in Table~\ref{table:baselines_7}, where the first three columns correspond to Greedy experiments and the last three to SA experiments. The Greedy and SA columns involve sampling new configurations only by moving atoms from occupied to empty positions, while the 70-30 columns sample new configurations with a 70\% probability to move an atom from a filled position to an empty one and a 30\% probability of swapping two atoms of different elements. The 50-50 columns select new configurations accordingly. We observe that for both the Greedy and the SA algorithms, the best configurations of every compound (except \ch{SrTiO3} with g=8 and Z=8) are obtained when choosing to move only atoms from occupied positions to empty ones. This happens because in the examined compounds the number of empty positions is much larger than the total number of atoms. As a result, the solution space is explored faster and more efficiently when atoms are allocated to previously empty positions.

\begin{table}[h!]
\begin{center}
\begin{tabular}{|c c c c c c c c|} 
 \hline
Structure & Metrics & Greedy & 50-50 & 70-30 & SA & 50-50 & 70-30
  \\
    \hline
  \ch{SrTiO3}, 
 g=4 & Mean Best ROG & 0 & 0 & 0 & 0 & 0 & 0 \\
   Z=1& Best ROG & 0 & 0 & 0 & 0 & 0 & 0\\
 -158.76 eV & Best energy & -158.76 & -158.76 & -158.76 & -158.76 & -158.76 & -158.76 \\
    \hline
  \ch{SrTiO3}, 
 g=8 & Mean Best ROG & 0 & 0 & 0 & 0 & 0 & 0 \\
   Z=1& Best ROG & 0 & 0 & 0 & 0 & 0 & 0\\
 -158.76 eV & Best energy & -158.76 & -158.76 & -158.76 & -158.76 & -158.76 & -158.76 \\
  \hline
  \ch{SrTiO3}, 
 g=8 & Mean Best ROG & 0.13 & 0.13 & 0.13 & 0.13 & 0.13 & 0.13 \\
 Z=8 & Best ROG & 0.12 & 0.11 & 0.12 & 0.11 & 0.11 & 0.11 \\
 -1268.67 eV & Best energy & -1121.39 & -1125.73 & -1118.1 & -1125.61 & -1127.18 & -1129.15\\
  \hline
 \ch{MgAl2O4}
 & Mean Best ROG & 0.10 & 0.10 & 0.10 & 0.10 & 0.10 & 0.10  \\
 & Best ROG & 0.09 & 0.09 & 0.09 & 0.09 & 0.10 & 0.09 \\
 -1620.89 eV & Best energy & -1483.11 & -1474.55 & -1476.47 & -1477.50 & -1465.88 & -1476.86 \\
    \hline
  \ch{Y2O3} 
  & Mean Best ROG & $0.16$ & 0.17 & 0.17 & $0.13$ & 0.18 & 0.17 \\
  & Best ROG & $0.15$ & 0.16 & 0.16 & $0.08$ & 0.16 & 0.15 \\
 -2191.57 eV & Best energy & -1869.21 & -1841.20 & -1843.13 & -1932.94 & -1833.96 & -1871.66 \\
  \hline
 \ch{Y2Ti2O7}
  &  Mean Best ROG & 0.17 & 0.17 & 0.16 & 0.16 & 0.16 & 0.16 \\
  &  Best ROG & 0.14 & 0.15 & 0.15 & 0.13  & 0.15 & 0.15 \\
 -3093.53 eV  & Best energy & -2660.99 & -2634.71 & -2637.75 & -2706.28 & -2631.52 & -2642.48 \\
    \hline
  \ch{SrTiO3},  
  g=8 & Mean Best ROG & 0.17 & 0.17 & 0.17 & 0.18 & 0.18 & 0.18 \\
 Z=27 & Best ROG & 0.16 & 0.16 & 0.16 & 0.16 & 0.17 & 0.17\\
 -4281.76 eV & Best energy & -3609.5 & -3588.22 & -3580.96 & -3600.31 & -3558.41 & -3551.13 \\
  \hline
 \ch{Ca3Al2Si3O12}
 & Mean Best ROG & 0.17 & 0.18 & 0.17 & 0.18 & 0.19 & 0.18 \\
 & Best ROG & 0.16 & 0.17 & 0.17 & 0.16 & 0.18 & 0.17\\
 -2199.99 eV & Best energy & -1840.75 & -1813.50 & -1833.20 & -1852.75 & -1799.60 & -1831.05 \\
    \hline
 \end{tabular}
 \caption{ROG and energy values for Greedy and SA under different configuration sampling strategies.}
\label{table:baselines_7}
\end{center}
\end{table}

\clearpage

\end{document}